\documentclass{article}
\usepackage[utf8]{inputenc}
\usepackage[margin=1in]{geometry}

\usepackage{microtype}
\usepackage{graphicx}
\usepackage{subfigure}
\usepackage{booktabs} 
\usepackage{natbib}
\setcitestyle{open={(},close={)}}

\usepackage[colorlinks=true,citecolor=blue]{hyperref}

\usepackage{amsmath,amsthm,amsfonts,amssymb,mathdots,array,mathrsfs,bm,bbm,stmaryrd,graphicx,subfigure,xcolor}
\usepackage{algpseudocode,algorithm,algorithmicx}
\usepackage{breakcites}

\usepackage[T1]{fontenc}
\usepackage{enumerate}
\usepackage{inputenc}

\usepackage{graphicx} 
\usepackage{subfigure}

\usepackage{booktabs,balance}
\usepackage{rotating}
\usepackage{boldline}
\usepackage{makecell}
\usepackage{multirow}
\usepackage{balance}

\usepackage{tikz}

\newtheorem{theorem}{Theorem}[section]
\newtheorem{corollary}[theorem]{Corollary}

\newtheorem{definition}{Definition}[section]

\newcommand{\bsmat}{\begin{bmatrix} }
\newcommand{\esmat}{\end{bmatrix} }

\usepackage{environ}
\NewEnviron{smallequation}{%
    \begin{equation}
    \scalebox{0.97}{$\BODY$}
    \end{equation}
    }
    \NewEnviron{smallalign}{%
    \begin{equation}
    \scalebox{0.97}{$\BODY$}
    \end{equation}
    }

\newcommand{\setarm}{\mathcal{K}}
\newcommand{\numarm}{K}
\newcommand{\arm}{k}
\newcommand{\optarm}{i^*}
\newcommand{\expec}{\mathbb{E}}
\newcommand{\reward}{X}
\newcommand{\tme}{t}
\newcommand{\rewardkt}{X^{(\arm)}_\tme}
\newcommand{\rwdst}{F}
\newcommand{\pol}{I_t}
\newcommand{\rwdpol}{X^{(\pol)}_\tme}
\newcommand{\mxrwd}{\mathcal{M}}
\newcommand{\indx}{n}
\newcommand{\nrwdpol}{X^{(I_\indx)}_\indx}
\newcommand{\filtration}{(\mathcal{F}_0,\mathcal{F}_1,\cdots)}

\newcommand{\filtt}{\mathcal{F}_\tme}
\newcommand{\val}{V_\tme}

\newcommand{\dur}{T}
\newcommand{\narm}{N_k(t)}
\newcommand{\marm}{m(t)}
\newcommand{\ostat}{\mathcal{O}_{\arm,\tme}(\zeta)}
\newcommand{\postat}{\mathcal{O}_{\arm,\tme}}
\newcommand{\ipol}{W}
\newcommand{\ipolkt}{W_k(t)}
\newcommand{\stsz}{\varepsilon}
\newcommand{\apol}{\widehat{W}}
\newcommand{\setmin}{\mathcal{I}}
\newcommand{\ami}{\xi}
\newcommand{\rv}{J_\tme}

\newcommand{\one}{\mathbf{1}}

\newcommand{\E}{\mathbb{E}}

\newcommand{\bma}{\begin{matrix*}[r]}
\newcommand{\ema}{\end{matrix*}}

\newcommand{\vep}{\varepsilon}

\DeclareMathOperator*{\argmax}{arg\,max}

\begin{document}

\title{\bf Extreme Bandits using Robust Statistics}

\author{\textbf{Sujay Bhatt,  Ping Li} \\
Cognitive Computing Lab\\
Baidu Research\\
10900 NE 8th St. Bellevue, WA 98004, USA\\
  \texttt{\{sujaybhatt,\ liping11\}@baidu.com}\\
 \and
 {\bf Gennady Samorodnitsky} \\
 School of ORIE\\
 Cornell University\\
220 Frank T Rhodes Hall, Ithaca,  NY 14853, USA\\
\texttt{gs18@cornell.edu}
}
\date{\vspace{0.5in}}

\maketitle

\begin{abstract}\vspace{0.1in}
\noindent\footnote{The work of Gennady Samorodnitsky was conducted as a consulting researcher at Baidu Research -- Bellevue, WA.}We consider a multi-armed bandit problem motivated by situations where only the extreme values, as opposed to expected values in the classical bandit setting, are of interest. We propose distribution free algorithms using robust statistics and characterize the statistical properties. We show that the provided algorithms achieve vanishing extremal regret under weaker conditions than existing algorithms. Performance of the algorithms is demonstrated for the finite-sample setting using numerical experiments. The results show superior performance of the proposed algorithms compared to the well known algorithms.
\end{abstract}

\newpage

\section{Introduction}
Multi-armed bandit (MAB) is a sequential decision making framework that formalizes the explore \& exploit trade-off under uncertainty. Here, the goal is to devise active sampling algorithms to identify sources generating the largest cumulative payoff~\citep{LR85,BC12,BCL13,Sli19,LS20}. In this paper, we focus on a  special class of MAB's called the \textit{Extreme Bandits}. 
Extreme Bandits or Max-K Bandits~\citep{SS06a,SS06b} are motivated by situations where only the extreme values~\citep{SHS05,CS05}, as opposed to total expected rewards in the classical bandit setting, are of interest.  In~\cite{NLB16}, it is shown that subtleties arise in the extreme bandit setting that are absent in the standard bandit setting. Using a more general regret definition, they show that no policy can be guaranteed to perform asymptotically as well as an oracle that plays the single best arm over a given duration. Thus this subclass merits independent attention owing to peculiarities observed only in the extreme bandit setting.

\vspace{0.1in}

\noindent
\textbf{Applications}: 
Real-world situations where non-parametric extreme-bandits algorithms are naturally useful have been described in literature. For example, the randomized search situations discussed in~\cite{CS05} can lead to either light-tailed or heavy-tailed reward distributions (though the paper itself utilizes a parametric approximation, the GEV distribution). The many different anomaly detection situations discussed in~\cite{CV14} and the references therein naturally include a large variety of reward distributions, some with light tails, and some with heavy tail. In solving NP-hard combinatorial optimization problems using stochastic search heuristics~\citep{CS05,SS06a}, where the current reward is the best solution found so far, the goal of future restarts is to find a solution that is better than the current best found. 

Extreme bandit setting is also applicable in many real-world problems in diverse fields such as telecommunications, epidemiology, molecular biology, astronomy, quality control, where the objective is to detect sources that behave normally most of the time, but sometimes experience a burst of extreme events~\citep{GPW09}; although in a limited bandit feedback setting. 
A complicated real-world situation is described in~\cite{AKAW04}, where it might be possible to decide on a specific distributional model for specific situations, and then design an extreme-bandit algorithm for that model. In long tail online marketing~\citep{SEH10}, for example, the marketer seeks to identify those markets that generate the largest sales traffic on individualized/niche products located in the long tail. These applications are naturally framed in the extreme bandit setting, which deals with sequentially choosing the distribution from a collection to sample in order to  maximize the single best reward.    

\vspace{0.1in}

\noindent
\textbf{Related Work}: There are numerous algorithms in the literature for solving extreme bandits, and these can be broadly divided into three categories:~\textit{Parametric algorithms}, where the distributions of the rewards are assumed to belong to specific distributions, for example Gumbel or Fr\'echet~\citep{CS05,SS06b}.~\textit{Semi-parametric algorithms}, where weaker semi-parametric assumptions on the distributions of the rewards are assumed, for example second order Pareto family~\citep{CV14,ACGS17} or a known lower bound on the tail distribution~\citep{DS16}. The above parametric/ semi-parametric settings leads to the natural questions of robustness of the algorithm with respect to inevitable deviations from the model. A \textit{distribution-free} algorithm that is shown to be efficient in variety of situations, including both light-tailed distributions and heavy-tailed distributions, may avoid such issues. Another motivation behind designing extreme-bandit algorithms that do not make any parametric assumptions on the distributions of the reward is similar to the motivation behind classical exploration-vs-exploitation algorithms, such as the Upper Confidence Bound (UCB)-type of algorithms~\citep{LR85}, for the usual average reward bandits. Even though the analysis of such algorithms often requires assumptions on the reward distributions (for example, sub-Gaussianity), there is nothing inherently parametric in the algorithm, which realizes its objective whether or not the rewards have, say, approximately, normal or beta distributions. \cite{SS06a} provide a distribution free algorithm for extreme bandits that works well for bounded rewards from any distribution that satisfies certain tail properties. However, \textit{no analysis} of the algorithm is provided. 

\newpage

\noindent
\textbf{Main Contributions}: We provide a \textit{distribution-free} extreme bandit algorithm and analyze its statistical properties. It is a novel index based algorithm, where the index is constructed in a non-parametric way by considering maximum elements of carefully designed sub-sets of observed data and then computing the median of these extreme values. Instead of the optimism principle in ExtremeHunter~\citep{CV14}, we use a particularly constructed randomization that allows one to explore arms whose index is not currently the highest. We also  establish the asymptotic correctness of the algorithm. We further provide a mollified algorithm, having the same asymptotic properties, however, which is also effective in identifying best arms distinguished only by the scaling coefficients. Finally, we establish vanishing extremal regret in the strong sense~(see (\ref{eq:RSS})) for exponential-like and polynomial-like distributions under \textit{weaker} assumptions than the state-of-the-art algorithms. This implies that there is no asymptotic regret of not knowing the best arm ahead of time.

\section{Extreme Bandit Setting}
Let $\setarm = \{1,2,\cdots,\numarm \}$ denote the arms of the multi-armed bandit, where  each arm~$\arm \in \setarm$ is associated with a reward distribution~$\rwdst_\arm$ having a \textit{finite} mean. Informally, at each step one ``pulls an arm'' and
obtains an independent observation from the distribution corresponding to that arm. Let~$\rewardkt, \tme \in \mathbb{Z}^+$  be i.i.d random variables from the distribution~$\rwdst_\arm$ for $\arm \in \setarm$. Let~$\pol \in \setarm$ be the arm pulled at time~$\tme$ to receive a reward~$\rwdpol$. Denote the maximum reward obtained by time $\tme $ as~$\mxrwd_\tme = \max_{ \indx \leq \tme}~\nrwdpol$. Define a filtration~$\filtration$, where~$\mathcal{F}_0$ is the
trivial $\sigma-$field and $\filtt = \sigma(I_1,X^{(I_1)}_1, \cdots
I_\tme, \rwdpol)$. For a time horizon $t$ a bandit strategy~$\pi_t =
(I_1,I_2,\cdots,I_t),~t \in \mathbb{Z}^+$, where each~$I_n$ is
$\mathcal{F}_{n-1}$ measurable is a legitimate strategy. Let~$\Pi_t$
denote the collection of all legitimate strategies. The \textit{\textbf{goal}} in an
extreme bandit setting is to find~$\pi_t \in \Pi_t$ such that
\begin{align} \label{eq:Val}
V_t(\pi_t):=\expec_{\pi_t}\mxrwd_\tme
\end{align}
is as large as possible.
In general, the optimal policy may depend on~$\tme$. 
An \textit{oracle} who knows the distributions~$\rwdst_\arm,~\arm \in \setarm$, would have chosen a strategy~\[
\pi^*_\tme \in \argmax_{\pi_t \in \Pi_t} \val(\pi_t).\]

In contrast, the
classical multi-armed bandit problem aims to solve the problem similar
to maximizing~(\ref{eq:Val}) but with~$\mxrwd_\tme$ replaced by~$R_{\tme} = \sum_{j=1}^{\tme} X_{j}^{(I_j)}$
and the regret of a policy in the classical sense is defined as
\begin{equation*}
    \mathcal{R}_{\tme}(\pi_t) = t\max_{i \in \mathcal{K}}
    \mathbb{E}[X^{(i)}_1] - \mathbb{E}_{\pi_t}[R_{\tme}].
  \end{equation*}
 It is well known that there exist multiple policies with a regret of
 the order $\log t$, that is 
with a \textit{\textbf{vanishing average regret}}, in the sense that~$\mathcal{R}_{\tme}(\pi_t)/t\rightarrow 0$. 

In case of extreme bandits, regret of any policy~$\pi_\tme \in \Pi_t$ can be obtained by comparing~$\val(\pi_\tme)$ and $\val(\pi^*_\tme)$. 
We will consider the situation usually studied in the literature on extreme bandits, where the existence of an asymptotically dominating arm is~assumed. An \textit{asymptotically dominating} arm~$\optarm$ is defined as:
\begin{equation*}
    \liminf_{n \rightarrow \infty} \frac{\mathbb{E}[\max_{j = 1,2,\cdots,n} X_j^{(\optarm)}]}{\mathbb{E}[\max_{j = 1,2,\cdots,n} X_j^{(i)}]} > 1
\end{equation*}
for each~$i \neq \optarm$, and we will try to detect and pull this arm most of the time.

\vspace{0.1in}
\noindent\textbf{Vanishing Extremal Regret.}\ The following notions of regret are considered for performance evaluation of the algorithm. Suppose~$\optarm$ is the asymptotically dominating arm. 
\begin{enumerate}
\item Vanishing extremal regret in a \textit{weak} sense:
\begin{align} \label{eq:RWS}
\frac{\expec_\pi [\max_{\indx  \leq \tme} \reward_\indx^{(I_\indx)}]}{\expec[\max_{\indx \leq \tme} \reward_\indx^{(\optarm)}]} \rightarrow 1,~\text{as}~\tme \rightarrow \infty.
\end{align}
\item Vanishing extremal regret in a \textit{strong} sense:
\begin{align} \label{eq:RSS}
\expec[\max_{\indx \leq \tme} \reward_\indx^{(\optarm)}] - \expec_\pi [\max_{\indx  \leq \tme} \reward_\indx^{(I_\indx)}]  \rightarrow 0,~\text{as}~\tme \rightarrow \infty.
\end{align} 
\end{enumerate}
Vanishing extremal regret is considered in~\cite{CS05,CV14,ACGS17}, with the aim of designing algorithms that detect an arm having the heaviest tail. This notion of regret is trivially achieved for distributions with bounded support for any policy
that chooses each distribution infinitely often. It provides a meaningful notion of regret with non-trivial policies for distributions with unbounded support~\citep{NLB16}. So we assume that the distributions~$F_k,~k \in \mathcal{K}$ have unbounded support with the only restriction of finite mean. 

\section{Max-Median Algorithm for Extreme Bandits} \label{Sec:MMA}
In this section, we provide a \textit{distribution-free} algorithm/ policy for extreme bandits. Without any room for confusion, we use policy and algorithm interchangeably. The algorithm named \textit{Max-Median}, is index based, whereby the index can be computed in~$O(KT \log T)$ time. 


\begin{algorithm}[h!]
\caption{Max-Median:\  $\text{MM}(\stsz_\tme, \{\nrwdpol\}$,  for $j \in \mathcal{K}$ and $n \leq t)$ }
 \label{Alg:MM}
\begin{algorithmic}[1]
\State $\tme-$run-time index. $\numarm-$number of arms. $\stsz_\tme-$decreasing step-size s.t $\sum \stsz_\tme = \infty$.  
\State $\pol \in \setarm-$arm chosen at~$\tme$.\  $\dur-$play horizon.\ 
$\narm-$number of $\arm^{th}$ arm pulls up to~$\tme$.
\State $\marm= \min_{k\in \setarm} \narm-$ minimum no. of pulls.
\State $\ostat-\zeta^{th}$ order statistic associated with the rewards from arm $k$. 
\State \textbf{Initialize}: Pull each arm once
\For{t = K+1: T}
\For{k = 1: K}
\State $\ipolkt := \mathcal{O}_{\arm,\tme-1}\Big(\Big\lceil \frac{  N_k(t-1)}{m(t-1)} \Big\rceil \Big)$    
\EndFor
\State $\pol = \Big\{\begin{array}{ll}
        \argmax_{\arm \in \setarm}~\text{\ipol}_\arm(\tme), &\text{w.p.}~1 -  \stsz_\tme\\
        i,\   \text{for } i \in \setarm &\text{w.p.}~ \frac{ \stsz_\tme}{\numarm}
        \end{array} $        
\EndFor
\end{algorithmic}
\end{algorithm}  

\textit{Discussion of~Algorithm~\ref{Alg:MM}:} Let~$\narm$ denote the number of times arm~$\arm$ is chosen up to time~$\tme$ with $\sum_k \narm = t$. Let
\begin{align} \label{eq:Mp}
\marm = \min_{k\in \setarm} \narm
\end{align}
denote the minimum number of times any arm is pulled. 

\begin{theorem} \label{thm:SS}
Let~$\stsz_\tme$ denote the decreasing step size such that $\sum
\stsz_\tme = \infty$. For any~$\ami > \numarm (\numarm-1)$,  $\marm$ is w.p.$1$ lower bounded for~$t$ large enough by~$\marm \geq \frac{1}{\ami}\sum_{d = 1}^{t} \stsz_d.$
\end{theorem}
Theorem~\ref{thm:SS} establishes that for $\xi>K(K-1)$, the event
$
\bigl\{ m(t)\geq (1/\xi)\sum_{d=1}^t
\epsilon_t \ \text{for all $t$ large enough} \bigr\}
$
has probability 1. This means that $m(t)$ will be greater than or equal to $(1/\xi)\sum_{d=1}^t \epsilon_t$ for all large $t$, but
the point from which this becomes true is still random, so for each {\it fixed} (large) $t$ the event  $\bigl\{ m(t)\geq (1/\xi)\sum_{d=1}^t \epsilon_t\bigr\}$ is a high probability event.\\

\noindent
\textit{Randomization}: The decreasing step size~$\stsz_\tme$ provides an avenue for exploration and hence plays a role in the rate of convergence of the extremal regret.  \\

\noindent
\textit{Index}: 
Let $H_\arm(\tme)$ be the set of times arm~$\arm$ is pulled by time
$\tme$.  Consider the following collection 
$\mathcal{S}_\arm(t) = \{A : A \subset  H_\arm(\tme), |A| =  m(t) \}$. 
It is clear that the cardinality  $|\mathcal{S}_\arm(t)| = \begin{pmatrix}
N_k(\tme) \\
m(\tme)
\end{pmatrix}$. Define~
$\widehat{\reward}_{\arm,j,\tme} = \max_{i\in A_j} X^{(\arm)}_i, \ j = 1,2, \cdots,  \begin{pmatrix}
N_k(\tme) \\
m(\tme)
\end{pmatrix}$ by enumerating the sets in $\mathcal{S}_\arm(t)$. We now introduce the following index:
\begin{align} \label{eq:Ain}
\apol_{\arm}(\tme) = \text{Median}~~\Big\{ \widehat{\reward}_{k,j,\tme-1},~j = 1,2, \cdots,  \begin{pmatrix}
N_k(\tme-1) \\
m(\tme-1)
\end{pmatrix}\Big\},
\end{align}
the median of all the maximum rewards on each subset of $m(\tme-1)$
rewards from arm $\arm$ observed before time $\tme$. These maxima are
available for each arm; they can be compared since they are taken over
sets of rewards of the same cardinality for each arm, and the  median is a
robust estimator of the size of these maxima, so it makes sense to
compare these medians. We argue that this index is close to the index
used in Algorithm~\ref{Alg:MM}. 
Let~$\ostat$ denote the $\zeta^{th}$ order statistic~\citep{Pic75,BC14} of the
$N_k(\tme)$ rewards from arm $k$ observed by time $t$; its computation
involves sorting, hence the complexity of $O(t\log t)$.  We have the following result.

\begin{theorem} \label{thm:Idx}
Let~$C>0$ be such that~$1 - x \geq e^{-Cx}$ for $0 \leq x \leq \frac{1}{2}$, and let $\tau = \frac{2\log (3/2)}{2C}$. For every arm~$\arm \in \setarm$, we have 
\begin{align*}
 \mathcal{O}_{\arm,\tme-1}\Big(\Big\lceil \frac{ 2
  N_k(t-1)}{m(t-1)} \Big\rceil \Big) \leq \apol_{\arm}(\tme)  \leq  \mathcal{O}_{\arm,\tme-1}\Big(\Big\lceil \frac{ \tau N_k(t-1)}{m(t-1)} \Big\rceil \Big),
\end{align*}
where $\lceil x \rceil = \min_{n \in \mathbb{Z}^+} \{n \geq x \}$, and
the upper bound holds  if $N_k(t-1)$ is large and $m(t-1) \leq N_k(t-1)/2$. 
\end{theorem}
Theorem~\ref{thm:Idx} is established using elementary combinatorics~\citep{Bru77}. Theorem~\ref{thm:Idx} suggests an index that is similar to the index in \eqref{eq:Ain} but computationally much simpler and easy to implement. This is summarized as follows.

\begin{corollary}
The index for the Max-Median algorithm  given as
\begin{align} \label{eq:OIdx}
\ipolkt :=   \mathcal{O}_{\arm,\tme-1}\Big(\Big\lceil \frac{
  N_k(t-1)}{m(t-1)} \Big\rceil \Big)
\end{align}
is analogous to \eqref{eq:Ain} but computationally simpler. 
\end{corollary}

\noindent
\textbf{Implementation Summary:}\ Starting with~$\ipol_k(1) =
0,~\forall \arm \in \setarm$, play all arms once. Pick the arm with the highest reward with probability~$1-\varepsilon_1$. For
each~$\tme>2$, maintain the number of times each of the $\numarm$ arms is played, and also the minimum number. Sort the rewards on each arm online and select the order statistic corresponding to the
index~$\ipolkt$ in (\ref{eq:OIdx}). With probability $1-\stsz_t$ pick the arm with the largest order statistic, while with probability
$\stsz_t$ explore a random arm.

\section{A Tale of Two Distributions} \label{Sec:TD}
Typical distributions considered in the extreme bandits  literature
have ``exponential-like'' tails such as the Gumbel Generalized Extreme Valued (GEV)
distribution or the exponential distribution~\citep{CS05,SS06b}, or
``sub-exponential/ heavy tails''
like the Frech\'et GEV distribution or the 
Pareto distribution~\citep{HW84,CV14,ACGS17}. Although the GEV
distributions are the limiting distribution of the
maxima of i.i.d random variables~\citep{FT28,DF07} and hence
considered in the extreme bandits literature, these distributions are often not even
an approximately accurate model of the payoff distributions
encountered in practice~\citep{SS06a}. So we consider rewards with 
more general
exponential-like tails~\citep{RP11} and polynomial-like tails~\citep{CV14,ACGS17}. 

We first establish the consistency of the index and the vanishing
extremal regret for both exponential-like tails and polynomial-like tails.
Then we provide a mollified index algorithm based on the Max-Median
idea that 
identifies best arms distinguished only by their scaling coefficients.   

\subsection{Exponential-like Arms} \label{sec:EA}

We will show that Algorithm~\ref{Alg:MM} achieves vanishing extremal regret in the strong sense~(\ref{eq:RSS}) for exponential-like arms. The assumption of exponential-like
arms means that 
\begin{align} \label{eq:TE}
\bar{\rwdst}_\arm(x) = 1- \rwdst_\arm(x) \sim a_\arm e^{-\lambda_\arm x},~\arm \in \setarm
\end{align}
for some~$a_{\arm} >0$ and $\lambda_{\arm} > 0$. It is easy{\footnote{An even stronger statement for the
expectation~(\ref{eq:ExE}) that involves the coefficients~$a_\arm$ as
well is provided in the appendix section.}} to see that for each~$\arm$
\begin{align} \label{eq:ExE}
\expec[\max_{\indx \leq \tme} \reward_\indx^{(\arm)}] \sim \lambda_\arm^{-1} \log t,~t \rightarrow \infty.
\end{align}

\noindent
The best arm~$\optarm \in \setarm$ is identified by $0 <
\lambda_{\optarm} < \min_{\arm \neq \optarm} \lambda_\arm$.  
It follows from \eqref{eq:ExE} that for $k\not= i^*$, 
\begin{align} \label{eq:EBa}
\lim_{\tme \rightarrow \infty} \frac{\expec[\max_{\indx \leq \tme}
  \reward_\indx^{(\optarm)}]}{\expec[\max_{\indx \leq \tme}
  \reward_\indx^{(\arm)}]} > 1. 
\end{align}
 
\subsubsection{Index Consistency} 
\noindent
\begin{theorem} \label{thm:EThm1}
Assume that for all~$\delta>0$,
\[
\sum_{t=1}^\infty \exp\left\{ -\left( \sum_{n=1}^t \epsilon_n\right)^\delta\right\}<\infty.
\]
Let $\optarm$ denote the best arm as in (\ref{eq:EBa}). For the Max-Median policy (Algorithm~\ref{Alg:MM}), the following holds:
\begin{align*}
\mathbb{P}\bigl(\ipol_{\optarm}(\tme) > \ipolkt \ \text{for all} \
  \arm \neq \optarm \ \text{and all} \ \tme~\text{large enough}\bigr) = 1.
\end{align*}
\end{theorem}

\vspace{0.1in}
In other words, w.p.1  the best arm will have the largest index eventually. This is crucial to establish vanishing regret in case of both exponential and polynomial arms. This result essentially guarantees the asymptotic correctness of Algorithm~\ref{Alg:MM}.

\begin{theorem} \label{thm:EVER}
Let $\optarm$ denote the best arm as in (\ref{eq:EBa}).  For the Max-Median policy (Algorithm~\ref{Alg:MM}), the following holds:
\begin{align*}
\lim_{\tme \rightarrow \infty} \Big( \val(\pi) - \expec[\max_{\indx
  \leq \tme} \reward_\indx^{(\optarm)}] \Big) = 0, 
\end{align*}
where $ \val(\pi)$ is as in (\ref{eq:Val}).
\end{theorem}

\vspace{0.05in}

According to Theorem~\ref{thm:EVER}, when the distributions of the
rewards are exponential-like,  Algorithm~\ref{Alg:MM} achieves
vanishing extremal regret in the strong sense~(\ref{eq:RSS}). In other words, there is no asymptotic regret of not knowing the best arm ahead of time. 

\subsection{Polynomial-like Arms}
In this section, we show that Algorithm~\ref{Alg:MM} achieves
vanishing extremal regret in the weak sense~(\ref{eq:RWS}) for
polynomial-like arms. Under additional assumptions which are weaker
than the state-of-the-art algorithms, vanishing extremal regret in the strong
sense~(\ref{eq:RSS}) is achieved for polynomial-like arms as well. The
assumption of polynomial-like arms means that 
\begin{align*}
\bar{\rwdst}_\arm(x) = 1- \rwdst_\arm(x) \sim a_{\arm} x
  ^{-\lambda_{\arm}},~\arm \in \setarm 
\end{align*}
for some~$a_{\arm} >0$ and $\lambda_{\arm} > 1$. It is easy to see that
\begin{align} \label{e:Emax.pol}
\expec[\max_{\indx \leq \tme} \reward_\indx^{(\arm)}] \sim a_{\arm}^{1/\lambda_{\arm}} \Gamma (1 - 1/\lambda_{\arm})\, t^{1/\lambda_k},~\tme \rightarrow \infty.
\end{align}
Here~$\Gamma(z) = \int_{0}^{\infty} e^{-x} x^{z-1} dx$ denotes the
Gamma function.  The best arm~$\optarm \in \setarm$ is identified by $0 <
\lambda_{\optarm} < \min_{\arm \neq \optarm} \lambda_\arm$.  
It follows from \eqref{e:Emax.pol}  that for $k\not= i^*$, 
\begin{align} \label{eq:EBa.p}
\lim_{\tme \rightarrow \infty} \frac{\expec[\max_{\indx \leq \tme}
  \reward_\indx^{(\optarm)}]}{\expec[\max_{\indx \leq \tme}
  \reward_\indx^{(\arm)}]} =\infty. 
\end{align}

\subsubsection{Index Consistency}

We note that one can switch from exponential-like arms
to polynomial-like arms by exponentiating the former, and switch
back by taking the logarithm of the latter.  Since the statement of 
Theorem~\ref{thm:EThm1} is invariant under monotone transformation of
the rewards, the theorem holds for polynomial-like arms as well.

\begin{theorem} \label{thm:PVR}
Let~$\optarm$ denote the best arm as in (\ref{eq:EBa.p}).  For the
Max-Median policy (Algorithm~\ref{Alg:MM}), the following holds: 
\begin{align*}
\lim_{\tme \rightarrow \infty} \frac{\val(\pi)}{\expec[\max_{\indx
  \leq \tme} \reward_\indx^{(\optarm)}]} = 1. 
\end{align*}
\end{theorem}
\noindent
According to Theorem~\ref{thm:PVR}, when the distributions of the
rewards are polynomial-like,  Algorithm~\ref{Alg:MM} achieves
vanishing extremal regret in the weak sense~(\ref{eq:RWS}). 
\noindent
We now compare the performance of Algorithm~\ref{Alg:MM} with
ExtremeHUNTER~\citep{CV14} \& ExtremeETC~\citep{ACGS17}, which are specifically designed for the
second order Pareto family defined by 
\begin{align*}
|\bar{\rwdst}_\arm(x) - a_{\arm} x ^{-\lambda_{\arm}}| \leq c_{\arm} x
  ^{-\lambda_{\arm} (1 + \beta_{\arm})}, ~\arm \in \setarm, 
\end{align*}
where $\beta_1, \beta_2, \cdots, \beta_{K}$ and $c_1,c_2,\cdots,c_{K}$ are positive constants. ExtremeHUNTER/ ExtremeETC achieves (\ref{eq:RSS}) under the following condition~\citep{CV14,ACGS17}:
\begin{align} \label{eq:CV_cmp}
\min(\beta_1, \beta_2, \cdots, \beta_{K}) >  1/\lambda_{\optarm}.
\end{align}
We prove that that Algorithm~\ref{Alg:MM} achieves vanishing extremal
regret in the strong sense (\ref{eq:RSS}) under a weaker assumption
$\beta_{i^*} > 1/ \lambda_{i^*}$. 

\begin{theorem} \label{thm:PSS}
Suppose~$\beta_{\optarm} > 1/\lambda_{\optarm}$. For $\alpha \in
(1/\lambda_{\optarm}, 1)$, let the exploration probabilities be chosen as $\epsilon_{\tme} = (1 + \tme)^{-\alpha}$. For the Max-Median policy (Algorithm~\ref{Alg:MM}), the following holds:
\begin{align*}
\lim_{\tme \rightarrow \infty} \Big( \val(\pi) - \expec[\max_{\indx \leq \tme} \reward_\indx^{(\optarm)}] \Big) = 0
\end{align*}
where $ \val(\pi)$ is as in (\ref{eq:Val}).
\end{theorem}

Theorem~\ref{thm:PSS} says that under additional
assumption~$\beta_{\optarm} > 1/\lambda_{\optarm}$, which is clearly
weaker than~(\ref{eq:CV_cmp}), Algorithm~\ref{Alg:MM} achieves
vanishing extremal regret in the strong sense~(\ref{eq:RSS}). In other words, there is no asymptotic regret of not knowing the best arm ahead
of time. \\

\noindent
\textbf{Remark:} ExtremeHUNTER~\citep{CV14} is an extreme bandit algorithm that is designed with semi-parametric assumptions on the distributions of the rewards. Specifically, assuming that the rewards are realized according to a second-order Pareto family, one uses an asymptotic approximation of the expectation of the maximum of these random variables. This approximation, along with plug-in estimates of the parameters appearing in the approximation, is used to compute an index. The estimates are computed optimistically, to account for uncertainty. The policy is not randomized, and the arm with the largest index is pulled.
In contrast, we do not assume that the reward distributions belong to any specific (semi)-parametric family. The index in Max-Median is constructed in a non-parametric way by considering maximum elements of carefully designed sub-sets of observed data and then computing the median of these extreme values. Instead of the optimism principle in ExtremeHUNTER, we use a particularly constructed randomization that allows one to explore arms whose index is not currently the highest.

\section{Mollified Max-Median Algorithm} \label{Sec:MMMA}
The Max-Median algorithm (Algorithm~\ref{Alg:MM}) has been showed to
be effective for both exponential-like and polynomial-like arms when
the best arm $i^*$ satisfies $0 <
\lambda_{\optarm} < \min_{\arm \neq \optarm} \lambda_\arm$. 
In this section, we propose a mollified Max-Median algorithm that
can distinguish effectively between several arms with the same optimal
value of $\lambda_{\optarm}$ but different values of the scaling
coefficient $a_k$. That is, we consider the situation 
\begin{align} \label{e:best.arm}
0 < \lambda_{\optarm_1} = \cdots =\lambda_{\optarm_j} < \min_{k \neq
  \{ \optarm_1, \optarm_2,\cdots,\optarm_j\}}
  \lambda_{\arm},~\text{and}~a_{\optarm_1} > \max_{l = 2,\cdots,j}
  a_{\optarm_l}, 
\end{align}
applicable to both exponential-like arms and polynomial-like arms. 


\noindent
\begin{algorithm}[h!]
\caption{mollified Max-Median: m-MM($\stsz_\tme$, $h(\cdot)$, $\{\nrwdpol\}$ for $j \in \mathcal{K}$ and $n \leq t$)}
 \label{Alg:MMM}
\begin{algorithmic}[1]
\State $\tme-$run-time index. $\numarm-$number of arms.  
\State $\stsz_\tme-$decreasing step-size s.t $\sum \stsz_\tme = \infty$.  
\State $\pol \in \setarm-$arm chosen at~$\tme$. $\dur-$play horizon.
\State $\narm-$number of $\arm^{th}$ arm pulls upto~$\tme$.
\State $\marm= \min_{k\in \setarm} \narm-$ minimum no. of pulls.
\State $h(\marm)-$index mollifier.
\State $\ostat-\zeta^{th}$ order statistic associated with the rewards
from arm $k$. 
\State \textbf{Initialize}: Pull each arm once
\For{t = K+1: T}
\For{k = 1: K}
\State $\widetilde{\ipol}_{\arm}(\tme) := \mathcal{O}_{\arm,\tme-1} \Big(\Big\lceil \frac{  N_k(t-1)}{h(m(t-1))} \Big\rceil \Big)$    
\EndFor
\State $\pol = \Big\{\begin{array}{lr}
        \argmax_{\arm \in \setarm}~ \widetilde{\ipol}_\arm(\tme), & \text{w.p}~1 -  \stsz_\tme\\
        i, & \text{for } i \in \setarm~\text{w.p}~ \frac{ \stsz_\tme}{\numarm}
        \end{array} $        
\EndFor
\end{algorithmic}
\end{algorithm}  

\noindent\textit{Discussion of Algorithm~\ref{Alg:MMM}.}\ 
The mollifier essentially provides a rationale to select a moderately higher order statistic for the index of each arm. The implementation is similar to Algorithm~\ref{Alg:MM}, except the minor modification in the index calculation. The time complexity is again~$O(\numarm T \log T)$.

\begin{definition}[Mollifier] \label{def:IM}
A mollifier is any increasing function $h: (0,\infty) \rightarrow (0,
\infty)$  
such that $h(x) \rightarrow \infty$ as $x \rightarrow \infty$ and
$h(x) = o( x/\log x)$. 
\end{definition}

Here the notation ~$f_1(x) = o(f_2(x)) $ means that $\lim_{x
  \rightarrow \infty}  f_1(x)/f_2(x) = 0.$ 
Theorem~\ref{thm:MAS} guarantees the asymptotic correctness of the mollified  algorithm (Aglorithm~\ref{Alg:MMM}).

\begin{theorem} \label{thm:MAS}
Assume that there is $\kappa>0$ such that
\begin{equation} \label{e:extra.cond}
\sum_{n=1}^{\infty}\left(  \sum_{d=1}^{n}\vep_{d}\right)^{-\kappa}<\infty.
\end{equation}
Let $\optarm=i_1^*$ denote the best arm as in \eqref{e:best.arm}, and
let $h$ be a mollifier.  Under Algorithm~\ref{Alg:MMM}, the following
holds for either exponential-like or polynomial-like arms. 
\begin{align*}
\mathbb{P}\bigl(\widetilde{\ipol}_{\optarm}(\tme) >
  \widetilde{\ipol}_{\arm}(\tme) \ \text{for all} \
  \arm \neq \optarm \ \text{and all} \ \tme~\text{large enough}
\bigr) = 1.
\end{align*}
\end{theorem}

\noindent
\textbf{Remark:} Results similar to Theorem~\ref{thm:EVER},
Theorem~\ref{thm:PVR}, and Theorem~\ref{thm:PSS} can be established
for the mollified Max-Median algorithm (Algorithm~\ref{Alg:MMM}) using
arguments similar to those used for Algorithm~\ref{Alg:MM}. 
In words, Algorithm~\ref{Alg:MMM} achieves vanishing extremal
regret in the strong sense~(\ref{eq:RSS}) in case of exponential-like and polynomial-like arms.

\section{Numerical Results} \label{Sec:NR}
 We know from Section~\ref{Sec:TD} that Algorithm~\ref{Alg:MM} and Algorithm~\ref{Alg:MMM} achieve vanishing extremal regret. So the focus of this section is to evaluate finite sample performance. In this section, we empirically evaluate Algorithm~\ref{Alg:MM} \&~\ref{Alg:MMM} on synthetic data.  

\vspace{0.1in}
\noindent\textbf{Performance Evaluation Discussion} 
\begin{enumerate}
    \item We employ two measures for evaluating the empirical finite sample performance: (I)~Extremal regret as in (\ref{eq:RSS}) in a non-asymptotic sense; (II)~Percentage of best~arm~pulls. The motivation for having another performance evaluation criterion stems from the fact that extremal regret is defined in an asymptotic sense, and we shall see that smaller extremal regret over a finite horizon need not reflect optimal play. Percentage of best arm pulls is a natural candidate for evaluation as the goal in extreme bandits can be seen as one of extreme value source identification.
    \item We consider 3~types of reward distributions: \textit{polynomial arms} for motivating heavy tailed data~\citep{BCL13}, \textit{exponential arms} for motivating exponential tailed data~\citep{RP11,KKM13}, and \textit{Gaussian arms} for motivating real valued data~\citep{Lat16}. These distributions are sufficiently diverse to cover the commonly encountered reward distributions in bandit applications.
    \item There are classical bandit algorithms like Robust-UCB~\citep{BCL13} that are designed for bandits with heavy tails. In~\cite{ACGS17}, it is demonstrated that, even though the objectives are completely different, Robust-UCB performs comparably to ExtremeHUNTER in terms of regret under stronger assumptions. Additionally, we compare the performance against non-heavy tailed distributions as well. So we only focus on comparison of the Max-Median algorithm against other extreme bandit algorithms. 
    \item Time complexity: The time complexity of the implementation of the three algorithms is as follows: Max-Median (Algorithm~\ref{Alg:MM}) has~$O(KT \log T)$, ExtremeHUNTER~\citep{CV14}  designed for second order Pareto family has~$O(T^2)$, and ThresholdAscent~\citep{SS06a}, which is distribution free,  has $O(KT)$. It is noted that a faster version ExtremeETC~\citep{ACGS17}, which has similar performance as ExtremeHUNTER, has~$O(\log^6T)$.
\end{enumerate}

\vspace{0.1in}
\noindent\textbf{Experimental Setup.}\  
In all simulations below, the hyper-parameters of ExtremeHUNTER are chosen as in~\cite{CV14}, and the hyper-parameters of ThresholdAscent are chosen as in~\cite{SS06a} with manual tuning to obtain the best performance for a given distribution. The only choice parameter in Algorithm~\ref{Alg:MM} is the step size that controls the exploration. All algorithms are evaluated over~$5000$ plays or arm pulls and values averaged over~$500$ trajectories, i.e., the expectation in~(\ref{eq:RSS}) is over~$500$ trajectories. The number of arms~$\numarm$ is chosen to be different for different distributions, as it is known that the algorithms' performance relative each to other is also affected by the number of bandit arms~\citep{KP14}.

\begin{figure}[h!]
\centering
    \mbox{
    \includegraphics[width=2.4in]{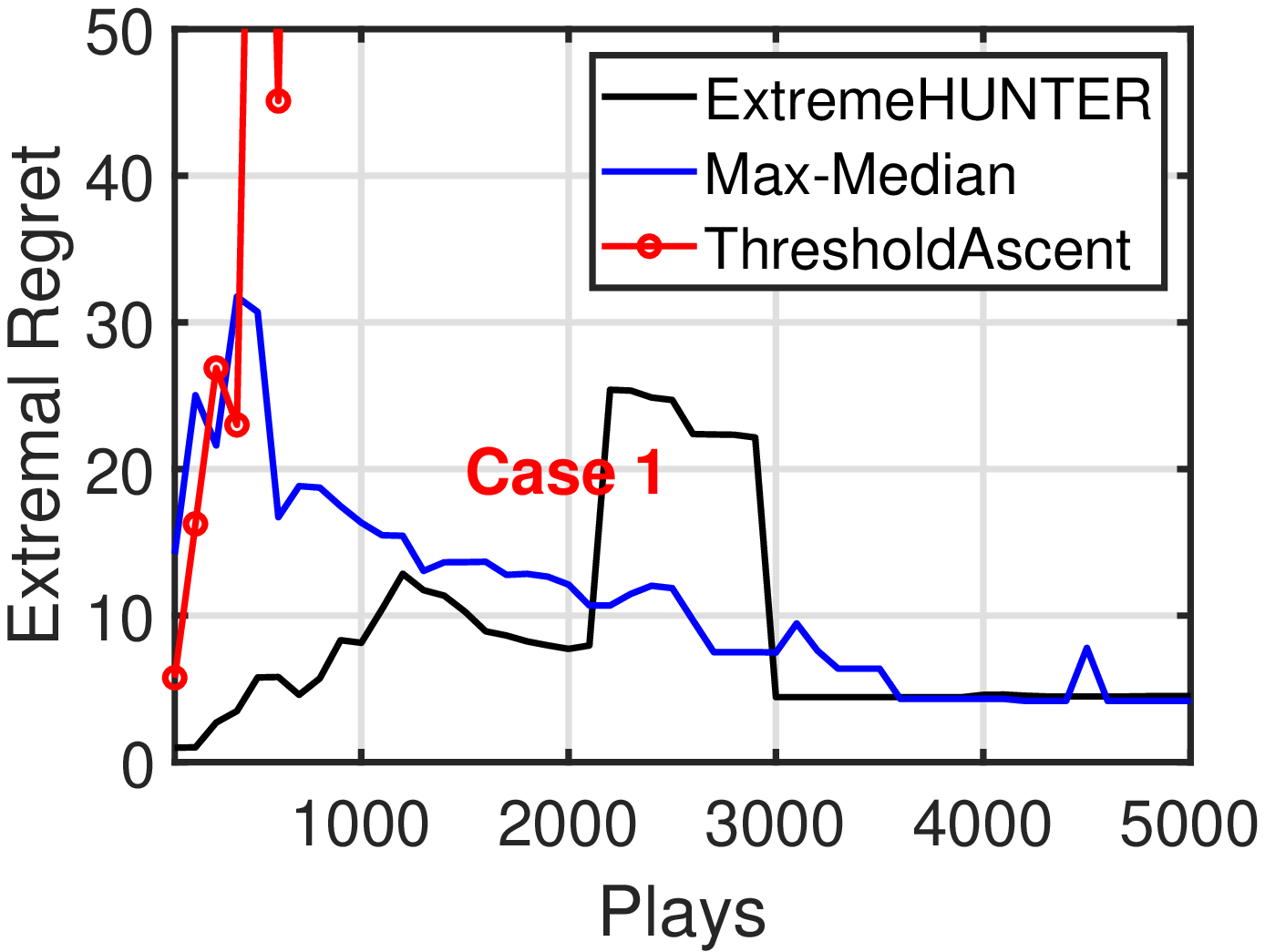}
    \hspace{0.3in}
    \includegraphics[width=2.4in]{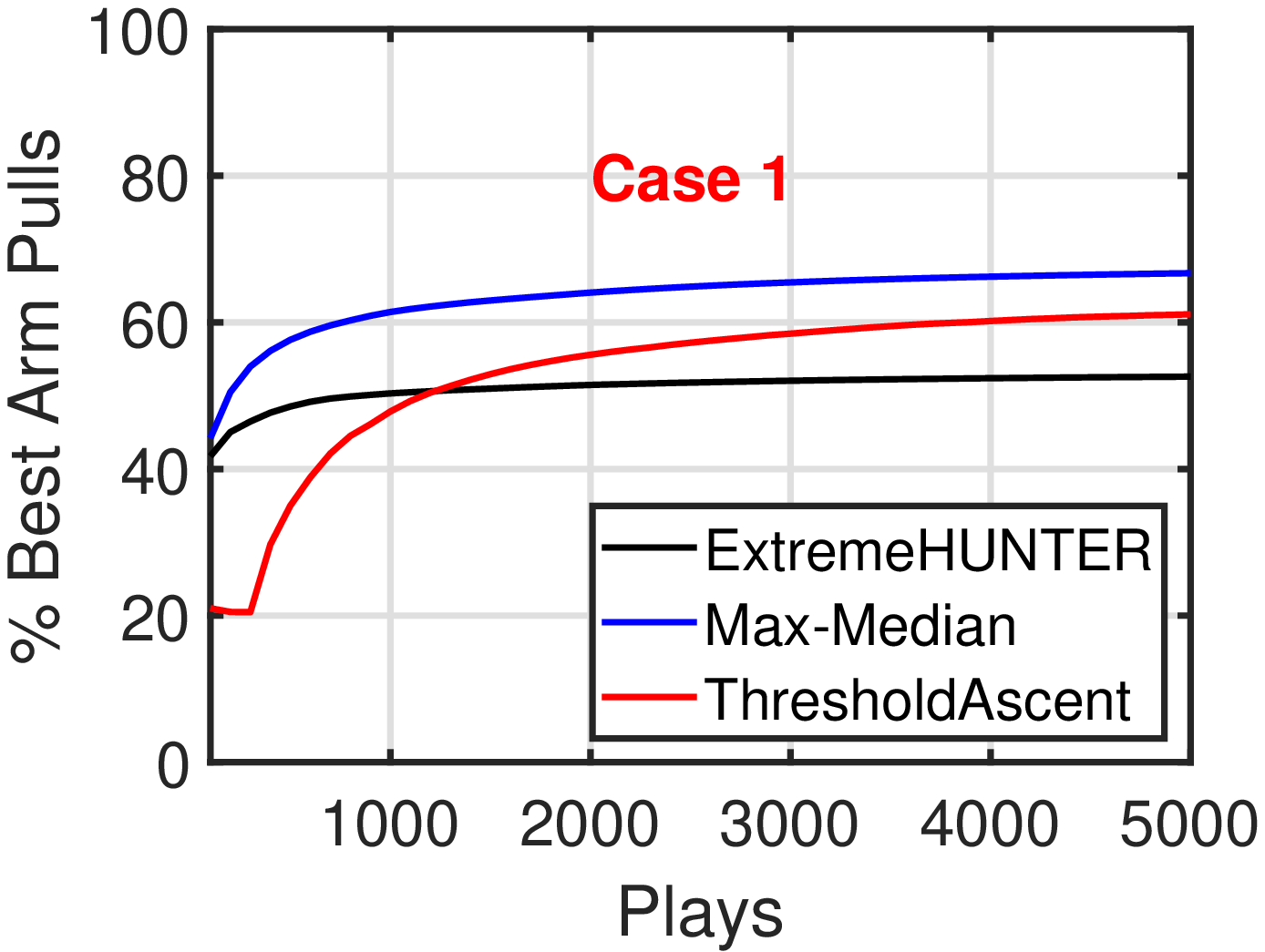}}
    
    \mbox{
    \includegraphics[width=2.4in]{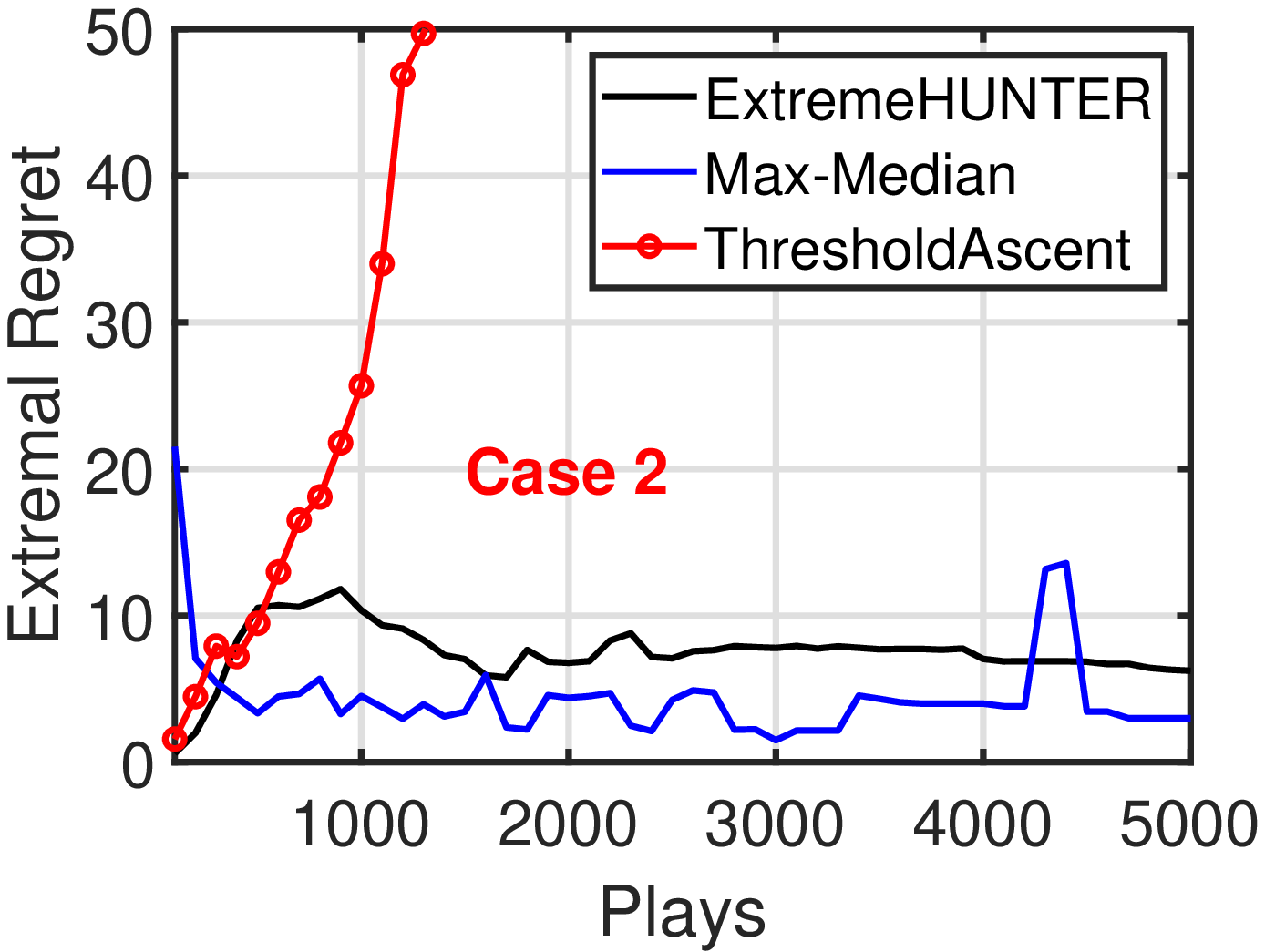}
    \hspace{0.3in}
    \includegraphics[width=2.4in]{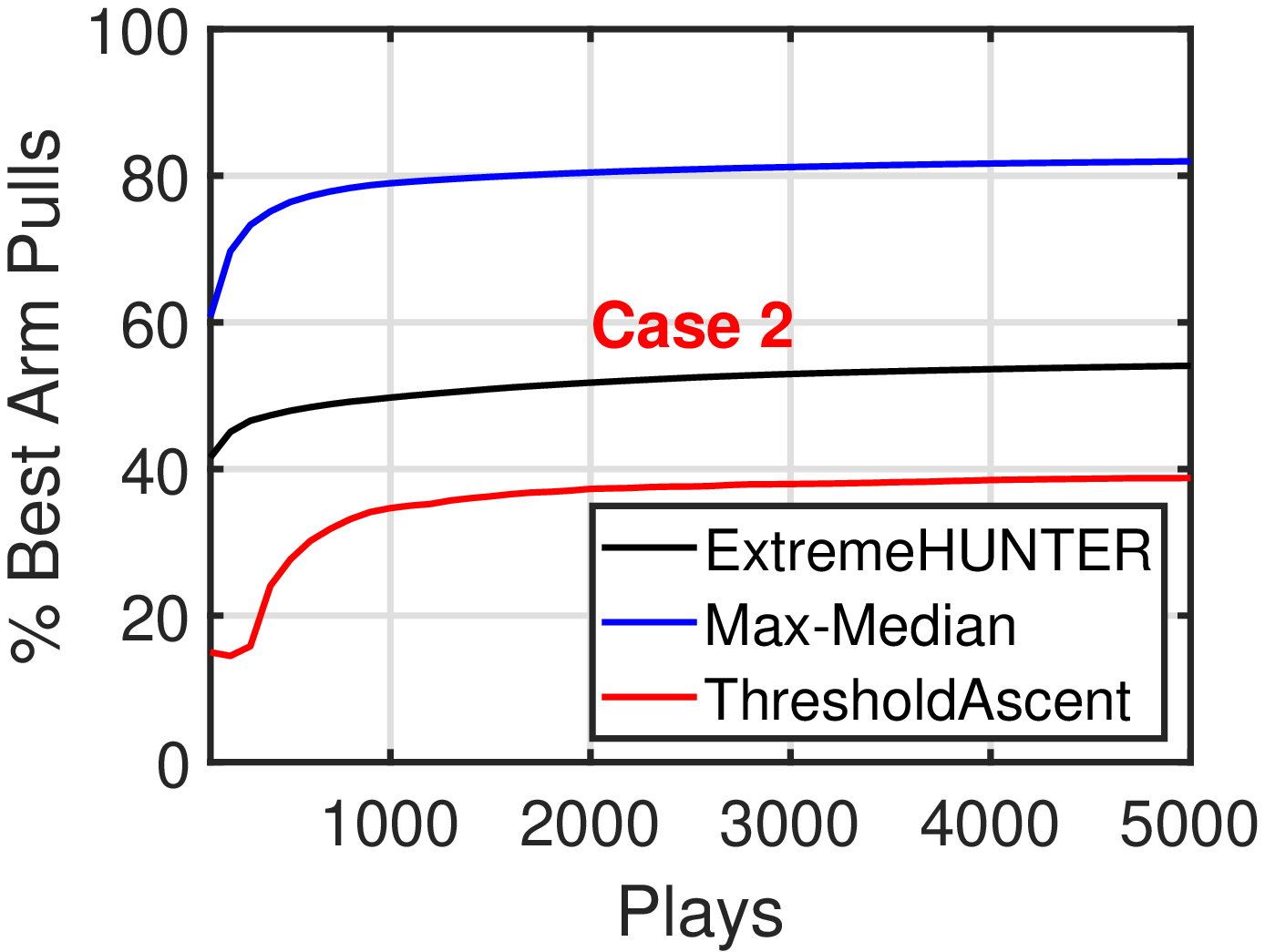}}
    
\vspace{-0.15in}
\caption{Finite sample performance for Polynomial Arms. ThresholdAscent performs poorly in terms of regret in both cases. Max-Median has similar regret performance as ExtremeHUNTER in both cases. Even though ExtremeHUNTER is specifically designed for Pareto distribution, it is dominated by Max-Median in both cases in terms of percentage of best arm pulled.} 
\label{fig:PDCases}\vspace{-0.1in}
\end{figure}

\begin{figure}[t!]     
\begin{center}
        \mbox{
    \includegraphics[width=2.4in]{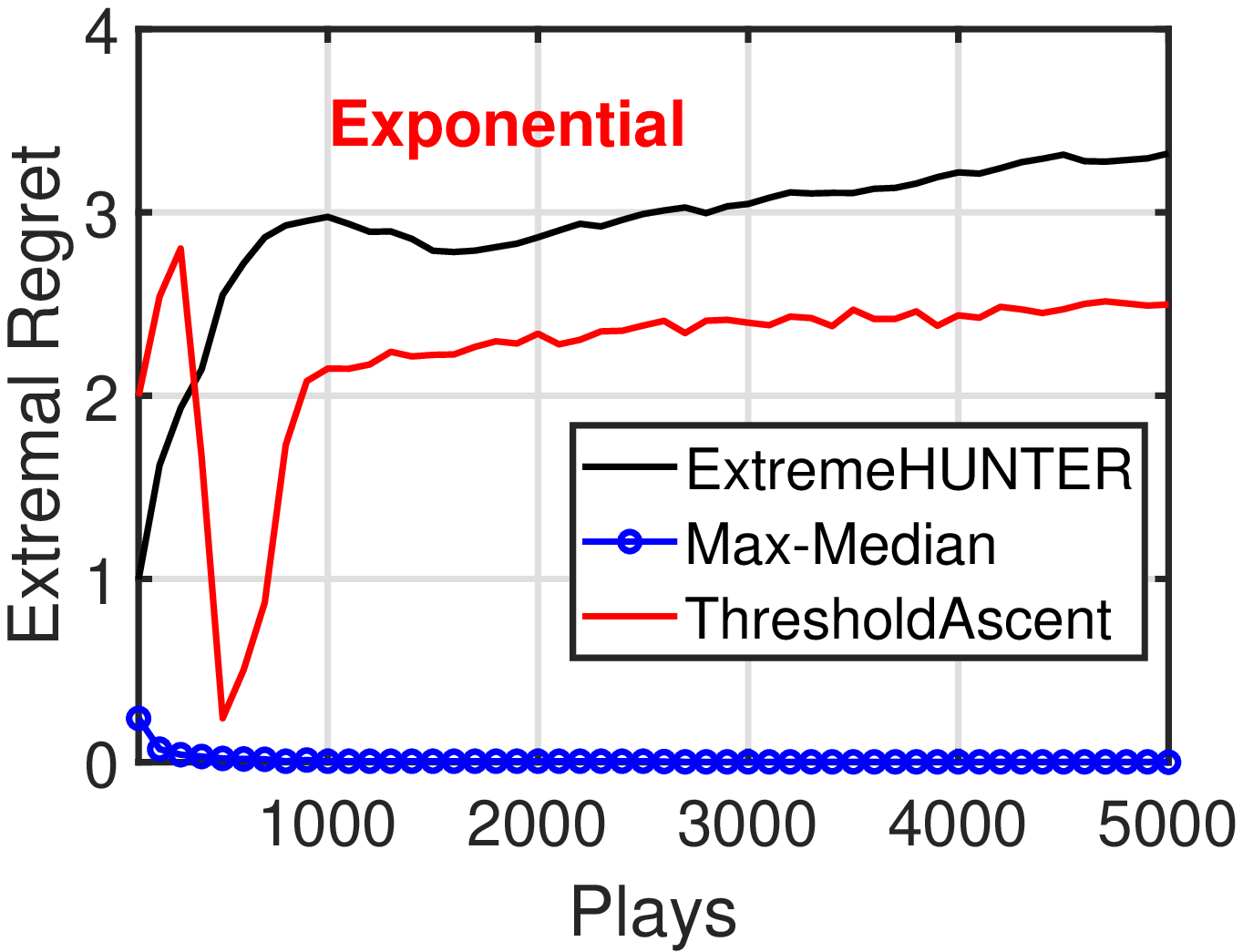}
    \hspace{0.3in}
    \includegraphics[width=2.4in]{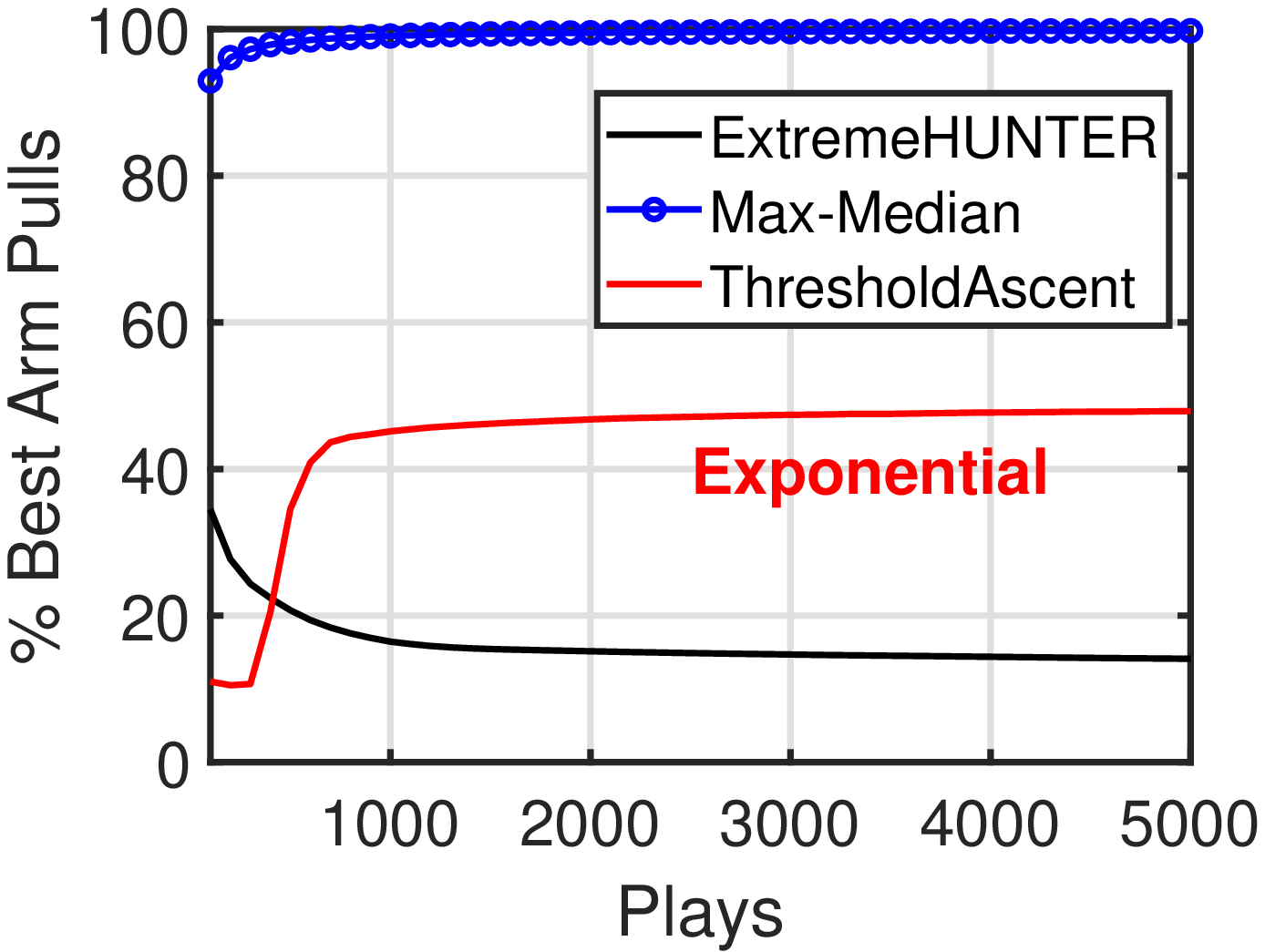}}
\end{center}
\vspace{-0.15in}
\caption{Finite sample performance for Exponential Arms. Max-Median performs extremely well for distributions with exponential tails. The best arm is identified in finite time while achieving finite time extremal regret. ThresholdAscent performs better than ExtremeHUNTER in  both evaluation criteria.} 
\label{fig:Exp}\vspace{-0.15in}
\end{figure}

\begin{figure}[t]     
\centering
    \mbox{
    \includegraphics[width=2.4in]{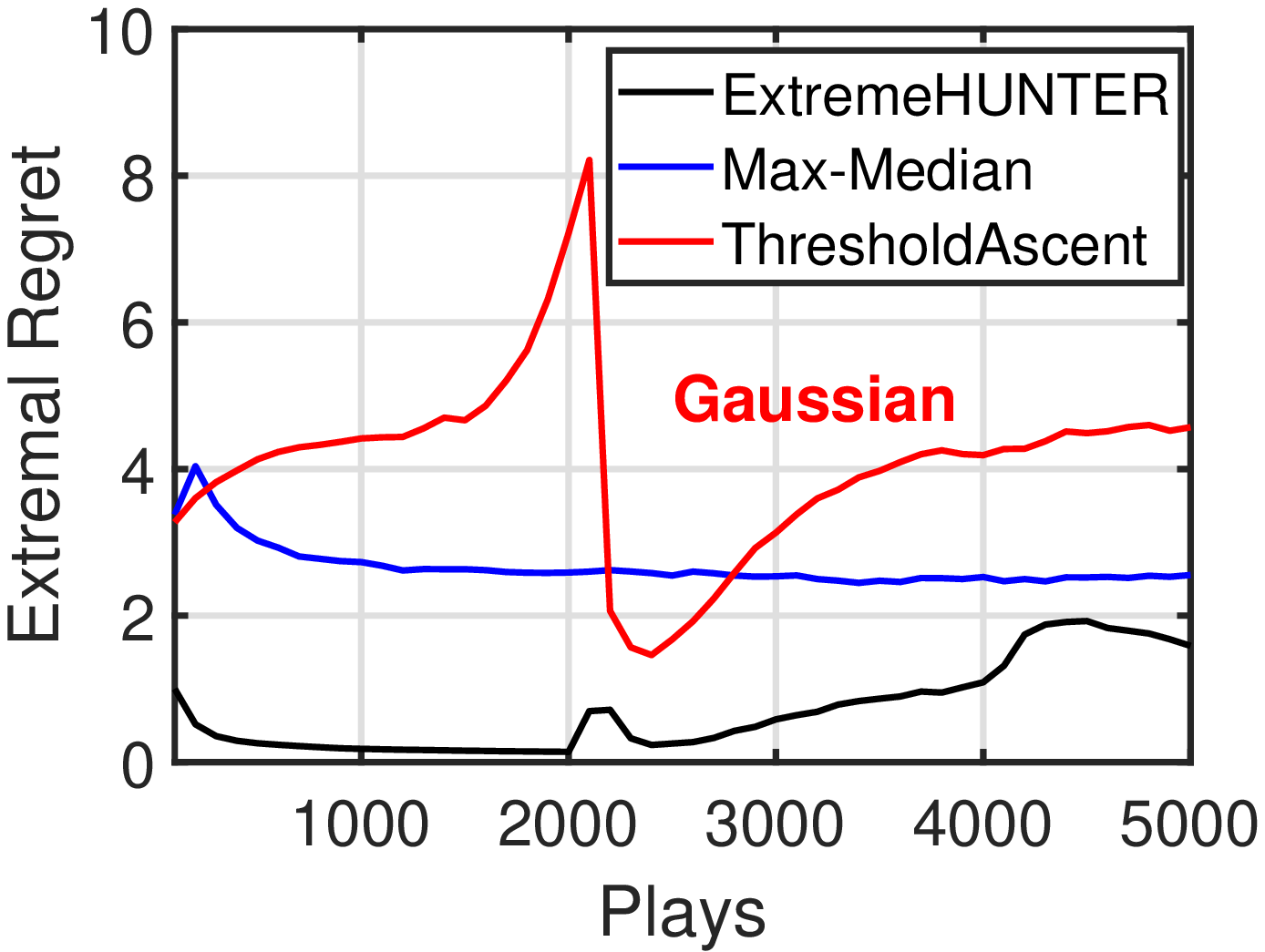}
    \hspace{0.3in}
    \includegraphics[width=2.4in]{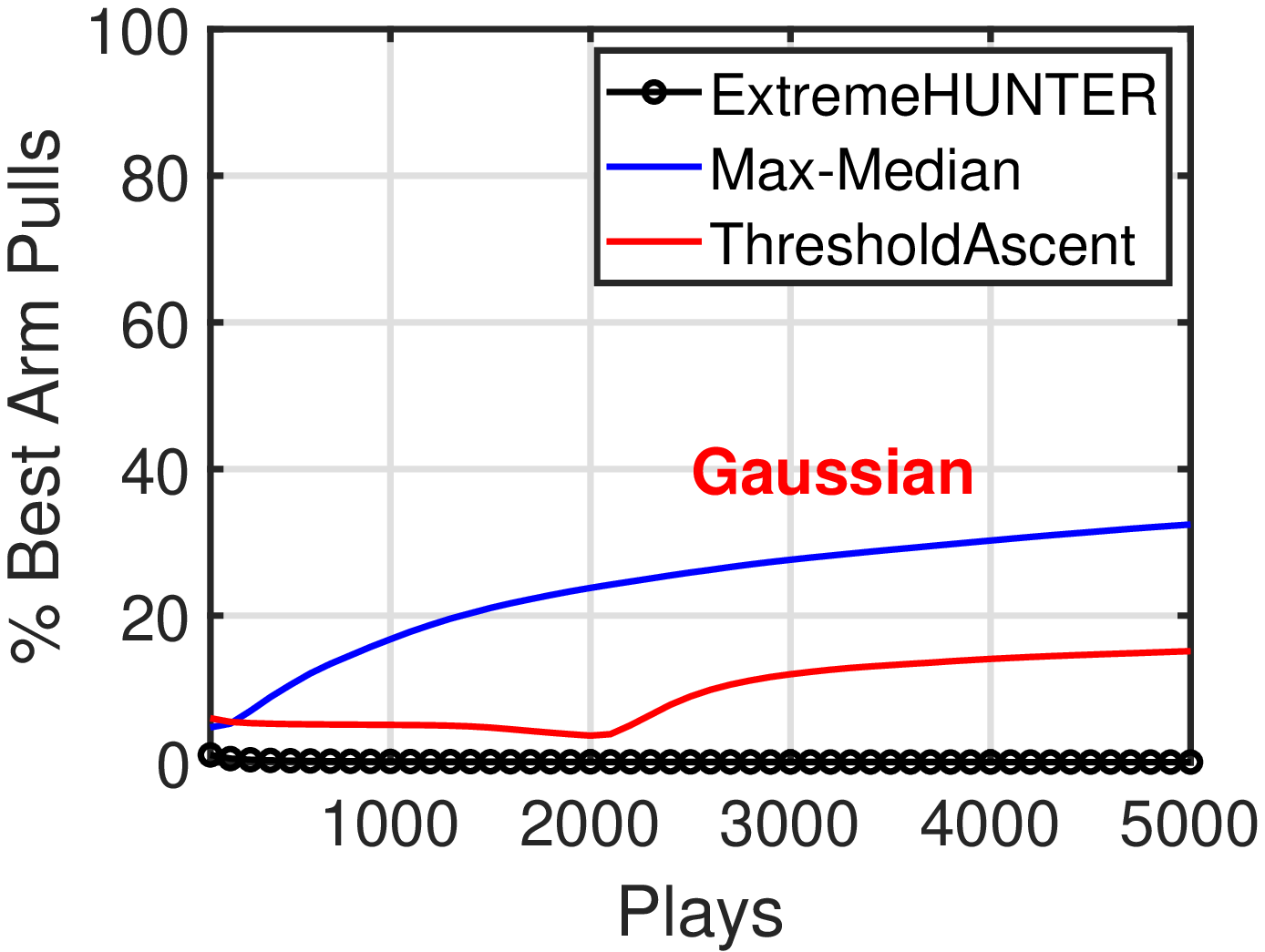}}
\vspace{-0.1in}    
\caption{Finite sample performance for Gaussian Arms. ExtremeHUNTER is unable to deal with Gaussian setting as the best arm is pulled less than~$3\%$ of the time. Max-Median is again the best performing algorithm in terms of percentage of best arm pulled. The dominating arm corresponds to the~one~with~the~largest~variance.} 
\label{fig:GauBer}
\end{figure}

\begin{enumerate}
\item \textbf{Polynomial Arms-Case 1}: We consider a~$K=5$ armed extreme bandit with polynomial arms having distinct distributional parameters~$\lambda_{\arm} = [2.1,2.3,1.3,1.1,1.9]$ and same coefficients. The distribution~$\bar{F}_\arm(x) \sim x^{-\lambda_\arm}$ is motivated by the numerical experiment in~\cite{CV14}, and is considered for fair comparison.  The step size that controls exploration in Algorithm~\ref{Alg:MM} is chosen as $\stsz_{\tme} = \frac{1}{(\tme+1)}$. The performance of the algorithms is illustrated in Figure~\ref{fig:PDCases}.
\item \textbf{Polynomial Arms-Case 2}: Next, we consider a related situation using a~$K=7$ armed extreme bandit with polynomial arms~$\bar{F}_\arm(x) \sim a_i x^{-\lambda_\arm}$ having similar distributional parameters for the best arm, where~$\lambda_{\arm} = [2.5,2.8,4,3.1,1.4,1.4,1.9]$ with $a_5 = 1.1, a_6 = 1.01$ and $a_{j\neq (5,6)} = 1$. The step size that controls exploration in Algorithm~\ref{Alg:MMM} is again chosen as $\stsz_{\tme} = \frac{1}{(\tme+1)}$ with the mollifier~$h(x) = \frac{\sqrt{x}}{\log x}$. This is equivalent to choosing a moderately higher-order statistic. The performance of the algorithms is illustrated in Figure~\ref{fig:PDCases}.

\newpage

\item \textbf{Exponential Arms}: Having considered a heavy tail setting with polynomial arms, we now consider an exponential tail setting with $\bar{F}_\arm(x) \sim e^{-\lambda_\arm}$. In this case, a $K=10$ armed exponential extreme bandit with $\lambda_{\arm} = [2.1,2.4,1.9,1.3,1.1,2.9,1.5,2.2,2.6,1.4]$ is considered. The step size in Algorithm~\ref{Alg:MM} is chosen as $\stsz_{\tme} = \frac{1}{(\tme+1)}$. The performance of the algorithms is illustrated in Figure~\ref{fig:Exp}.
\item \textbf{Gaussian Arms}: Motivated by applications having real valued extreme value source identification, we consider a Gaussian setting with~$K=20$ arms. For the purpose of illustrating the tail identification, we consider same means with different variances for the different arms, that is $\bar{f}_\arm(x) \sim \mathcal{N}(\mu_{\arm},\sigma_{\arm})$, $\mu_{\arm} = 1~\forall \arm \in \setarm$ and\footnote{$\sigma_{\arm} = [1.64, 2.29, 1.79,  2.67,  1.70,  1.36, 1.90, 2.19,   0.80, 0.12,  1.65,  1.19,  1.88,  0.89,  3.35, 1.5, 2.22,  3.03,   1.08,  0.48]$}. The step size in Algorithm~\ref{Alg:MM} is chosen as $\stsz_{\tme} = \frac{1}{(\tme+1)}$. The performance of the algorithms is illustrated in Figure~\ref{fig:GauBer}.
\end{enumerate}

\begin{figure}[h!]     
\begin{center}
    \mbox{
    \includegraphics[width=2.4in]{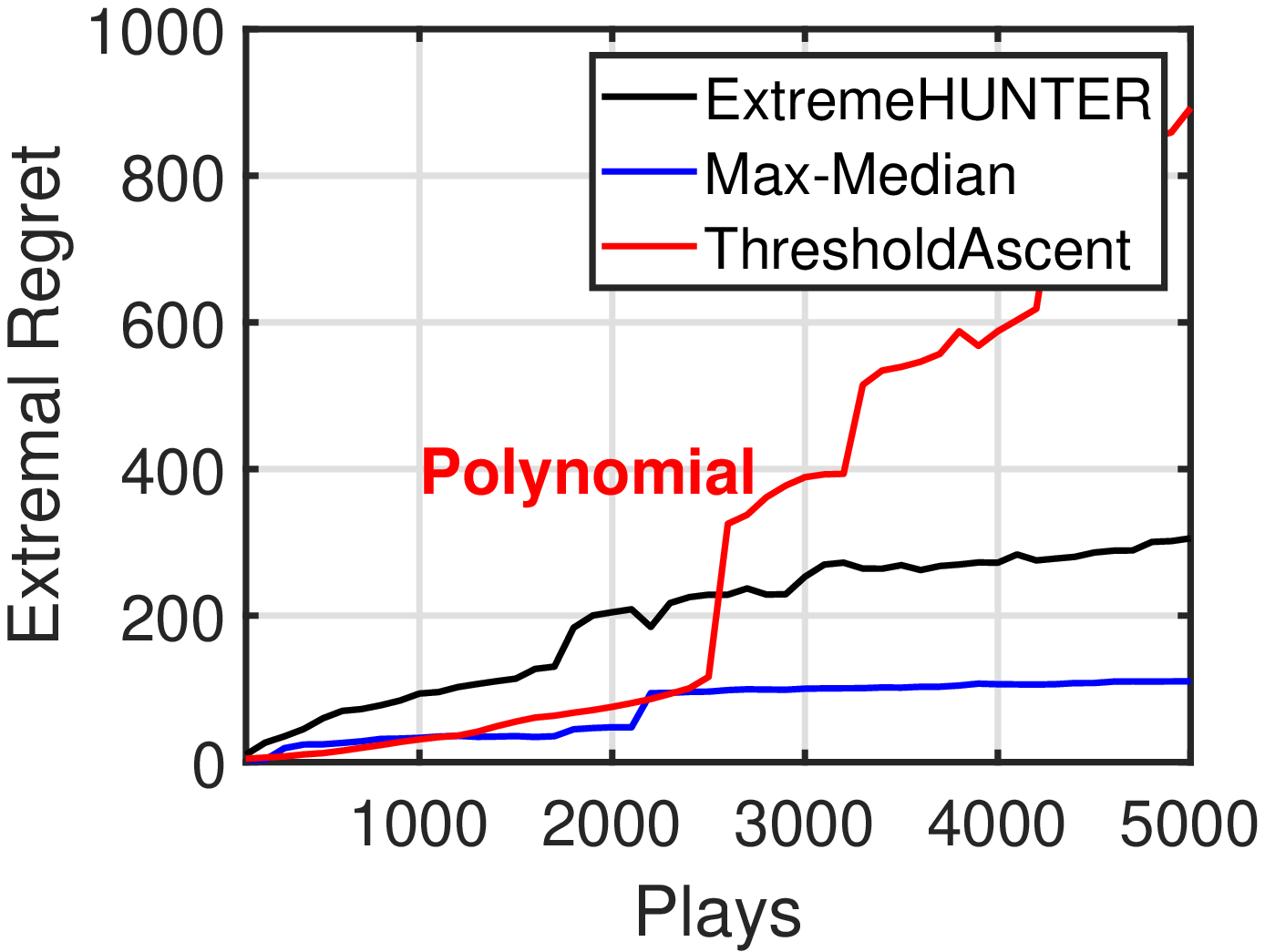}
       \hspace{0.3in}
    \includegraphics[width=2.4in]{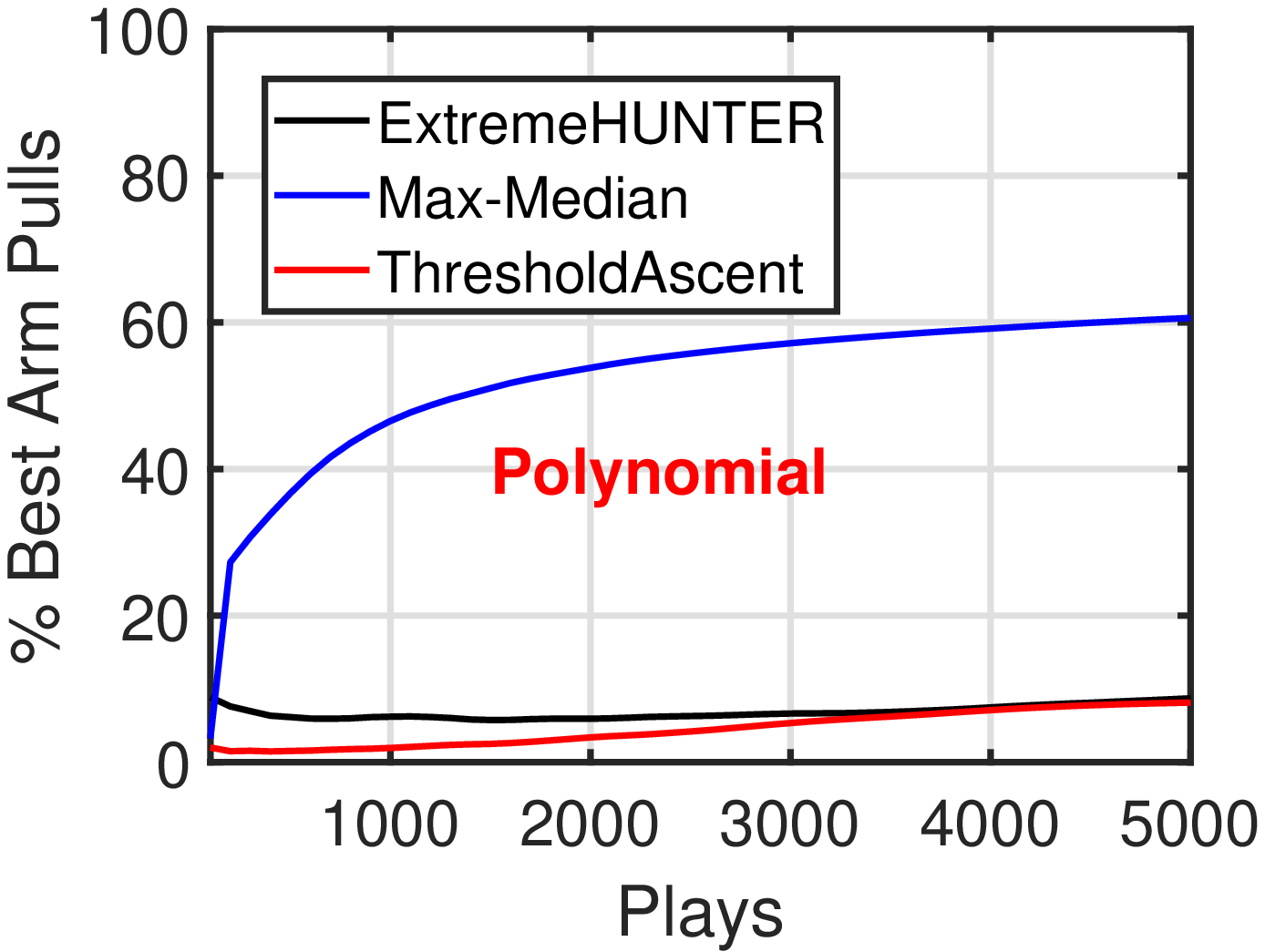}}
    \mbox{
    \includegraphics[width=2.4in]{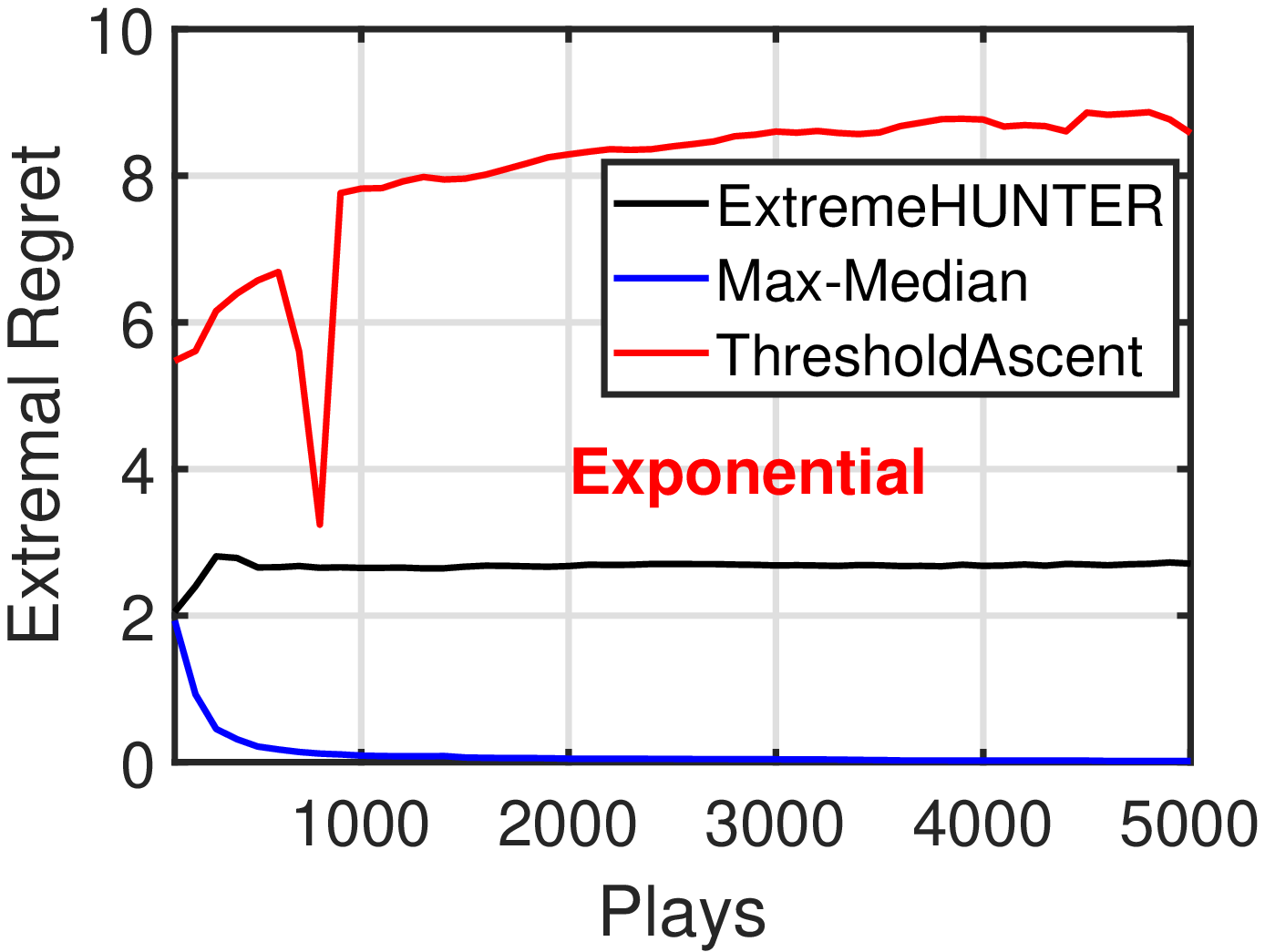}
       \hspace{0.3in}
    \includegraphics[width=2.4in]{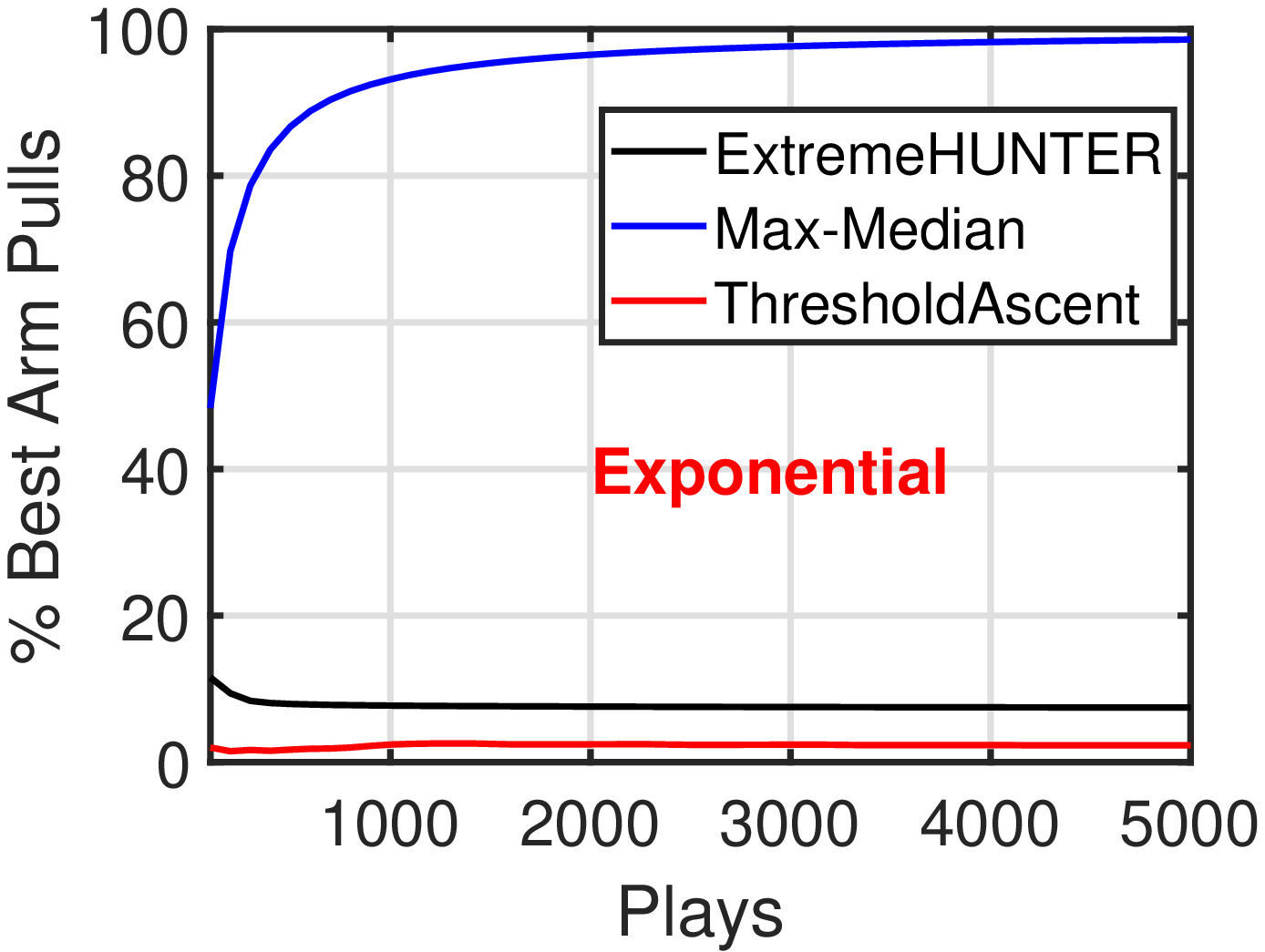}}
\end{center}
\vspace{-0.2in}
\caption{Finite sample performance for~$K=100$ Polynomial and Exponential Arms.} 
\label{fig:Pol}\vspace{-0.1in}
\end{figure}

\newpage

Next, we illustrate the performance of Algorithm~2 (Mollified Max-Median) in case of a large number of polynomial and exponential arms. The performance of Algorithm~2 in case of polynomial arms and exponential arms is shown in Figure~\ref{fig:Pol}. The mollifier is chosen as~$h(x) = \frac{\sqrt{x}}{\log x}$. The step size in Algorithm~2 is chosen as $\stsz_{\tme} = \frac{1}{(\tme+1)}$. The coefficients were chosen equal to~$1$ for all arms, and the lambda values were randomly generated using a power law distribution. It should be noted that for large number of arms, Algorithm~2 is preferred over Algorithm~1. 

\vspace{0.1in}
\noindent\textbf{Key Observations}
\begin{enumerate}
    \item Max-Median (Algorithm~\ref{Alg:MM}) has comparable extremal regret performance with ExtremeHUNTER, while performing the best amongst the chosen comparative algorithms in terms of percentage of best arm pulls, irrespective of the reward distribution. 
    \item ThresholdAscent performs poorly in case of all chosen distributions in terms of extremal regret~(\ref{eq:RSS}), while performing reasonably well in terms of percentage of best arm pulls. Even though ExtremeHUNTER performs well in terms of the regret, it performs poorly in all cases in terms of the percentage of best arm pulls. 
    \item From the empirical results, we infer that, for extreme bandits, extremal regret~(\ref{eq:RSS}) is not a good measure of performance over finite-horizon settings. The extremal regret being small does not reflect the fact the dominating arms are pulled most of the time.
    \item The good performance of Algorithm~\ref{Alg:MM} in finite sample settings for all chosen distributions, in terms of the percentage of best arm pulls, motivates the use of Algorithm~\ref{Alg:MM} for exploration in case of classical bandits~\citep{AB10,JMNB14}, and in combinatorial bandit problems~\citep{Ont13,NTGM18} for heavy-tail distributions. 
\end{enumerate}

\section{Conclusion}
We provided a general purpose algorithm for extreme bandits that has~$O(KT \log T)$ time complexity. The index based algorithm is fashioned using combinatorics and robust statistics, where we established that the index corresponding to the best arm will have the largest value asymptotically. We also provided a mollified algorithm to select the best arm, when only the distribution coefficients are distinct. Using numerical experiments, we demonstrated the superior finite-sample performance of the algorithm against the popular algorithms. Finally, to provide a comparison with the existing semi-parametric algorithms, we established vanishing extremal regret for the Max-Median algorithm for distributions having ``exponential-like tails'' and ``polynomial-like tails''- the most common class of distributions considered in the literature on extreme bandits- and demonstrated vanishing extremal regret under weaker conditions. It is likely that our algorithm is efficient in other situations as well, however, that is in consideration for future work.

The Max-Median algorithm uses forced randomization; and this has both advantages and drawbacks, and the analysis is sometimes more transparent in the randomized case. It is not quite clear how to construct a non-randomized procedure (such as utilizing optimism-in-face-of uncertainty) for the extremes without making some distributional assumptions. We are exploring this issue for future work.




\newpage

\noindent\textbf{\LARGE \bf Appendix}
\appendix

\section{Preliminaries}
We list here several properties of exponential-like and polynomial-like distributions. These properties are repeatedly used of in the
proofs of the theorems. 
\subsection{Exponential-like Arms}
Suppose the reward distribution of an arm is exponential-like: 
\begin{align*} 
\bar{\rwdst}(x) = 1- \rwdst(x) \sim a e^{-\lambda x} 
\end{align*}
for some~$a >0$ and $\lambda > 0$. Then a sample from this
distribution satisfies 
\begin{align*} 
\expec[\max_{\indx \leq \tme} \reward_\indx] \sim \lambda^{-1} \log
  t,~t \rightarrow \infty. 
\end{align*}
Moreover, 
\begin{equation}  \label{eq:ExE.ext}
 \lim_{\tme\to\infty} \E  \left[\max_{\indx \leq \tme}\reward_\indx- \lambda^{-1} \log
     \tme\right]= \lambda^{-1} \log a-
   \lambda^{-1}\int_0^\infty  e^{-x}\, \log x\, dx.
\end{equation}
\noindent
\begin{proof}[\textbf{Proof of \eqref{eq:ExE.ext}}]
  It is, clearly, enough to prove \eqref{eq:ExE.ext}  in the case
$\lambda=1$. Let $0<\vep<1$, and choose $M>0$ so large that both
$ae^{-M}\leq 1$ and 
$$
a^{-1}e^{x}P(X_1>x)\in [1-\vep,1+\vep]
$$
for all $x\geq M$. We have
\begin{align*} 
E\bigl[\max_{j=1,\ldots,n}X_j\bigr] =& \int_0^\infty \Bigl( 1-\bigl(
                           1-P(X_1>x)\bigr)^n\Bigr) dx \\
  =& \int_0^M\Bigl( 1-\bigl(
                           1-P(X_1>x)\bigr)^n\Bigr)dx +\int_M^\infty \Bigl( 1-\bigl(
                           1-P(X_1>x)\bigr)^n\Bigr)dx.\notag 
\end{align*}
It is clear that
\begin{equation} \label{e:det.1}
\lim_{n\to\infty} \int_0^M\Bigl( 1-\bigl(
1-P(X_1>x)\bigr)^n\Bigr)dx = M.
\end{equation}
Furthermore,
\begin{align*}
\int_M^\infty \Bigl( 1-\bigl(1-a(1-\vep)e^{-x}\bigr)^n\Bigr)dx \leq& 
\int_M^\infty \Bigl( 1-\bigl(1-P(X_1>x)\bigr)^n\Bigr)dx \\
\leq&  \int_M^\infty \Bigl( 1-\bigl(1-a(1+\vep)e^{-x}\bigr)^n\Bigr)dx.
\end{align*}
Write 
\begin{align*}
&\int_M^\infty \Bigl( 1-\bigl(1-ae^{-x}\bigr)^n\Bigr)dx
                 = \int_0^{ane^{-M}} \bigl( 1-(1-w/n)^n\bigr)\frac{dw}{w} \\
  =&\int_0^{ane^{-M}} \bigl( e^{-w}-(1-w/n)^n\bigr)\frac{dw}{w}
     + \int_0^{ane^{-M}} \bigl(1- e^{-w}\bigr)\frac{dw}{w}. 
\end{align*}
Changing the order of integration,
\begin{align*} 
 \int_0^{ane^{-M}} \bigl(1- e^{-w}\bigr)\frac{dw}{w}  
   =& (\log a + \log n -M)\int_0 ^{ane^{-M}} e^{-t}\, dt -
      \int_0^{ane^{-M}}   e^{-t}\, \log t\, dt \\
  \notag =
      &\log a + \log n -M - \int_0^\infty   e^{-t}\, \log t\, dt + o(1).
\end{align*}
Since we that 
\begin{align*}
0\leq  \int_0^{ane^{-M}} \bigl( e^{-w}-(1-w/n)^n\bigr)\frac{dw}{w} 
  \leq \int_0^{1} \bigl( e^{-w}-(1-w/n)^n\bigr)\frac{dw}{w} \to 0
\end{align*}
as $n\to\infty$, \eqref{eq:ExE.ext} follows. 
\end{proof}

\vspace{0.2in}

\noindent
Let $X_{[j:n]}$ be the $j$th largest order statistic from the sample
$X_1,\ldots, X_n$. Let $m_n\to\infty$, $m_n/n\to 0$ as
$n\to\infty$. Then for every $b>0$
\begin{equation} \label{e:ld.upper.exp}
  \limsup_{n\to\infty} \frac{m_n}{n}\log  P\left(  \big| X_{([n/m_n]:n)}
    -\lambda^{-1}\log a - \lambda^{-1} \log m_n\big|>b\right)  <0. 
\end{equation}

\begin{proof}[\textbf{Proof of \eqref{e:ld.upper.exp}}]
 Once again we may assume that $\lambda=1$. We have 
\begin{align*}
 P\left(  X_{([n/m_n]:n)}>
    \log a + \log m_n+b\right)  
= P\bigl( B_n\geq [n/m_n]\bigr),  
\end{align*}
where $B_n$ has the Binomial distribution with $n$ trials and
probability for success $p_n\sim e^{-b}/m_n$. By the exponential Markov
inequality, for any $\gamma>0$,
\begin{align*}
P\bigl( B_n\geq [n/m_n]\bigr) \leq \left[ e^{-\gamma [n//m_n]/n}\left(
  1+p_n\bigl( e^\gamma-1\bigr)\right)\right]^n.
\end{align*}
Choosing $\gamma=b$, we obtain
\begin{align*}
&\frac{m_n}{n} \log P\left(  X_{([n/m_n]:n)}>
                 \log a + \log m_n+b\right)  \\
  \leq &
 b \frac{ m_n}{n} -\bigl( b-m_np_n(e^b-1)\bigr) 
 \to -\bigl( b-1+e^{-b}\bigr)<0. 
\end{align*}
\end{proof}

\subsection{Polynomial-like Arms}
Suppose the reward distribution of an arm is polynomial-like: 
\begin{align*} 
\bar{\rwdst}(x) = 1- \rwdst(x) \sim a x ^{-\lambda}
\end{align*}
for some~$a >0$ and $\lambda > 1$. Then a sample from this
distribution satisfies 
\begin{align} \label{eq:SupPA}
\expec[\max_{\indx \leq \tme} \reward_\indx] \approx a^{ 1/\lambda}
  \Gamma (1 - 1/\lambda) t^{1/\lambda_k},~\tme \rightarrow \infty. 
\end{align}
Here~$\Gamma(z) = \int_{0}^{\infty} e^{-x} x^{z-1} dx$ for $z>0$ 
denotes the gamma function.

\section{Proofs of Main Results}
\noindent
\begin{proof} [\textbf{Proof of Theorem \ref{thm:SS}}]
From Algorithm~1, there is a sequence of independent $\{0,1\}$-valued random variables $\rv,~\tme=1,2,\ldots$ with $P(\rv=1)=\frac{\stsz}{(\numarm-1)}$, having the following property. For $\tme = 1, 2, \cdots$, let
\begin{equation*}
\setmin_\tme = \{\arm \in \setarm: \narm = \marm \}, 
\end{equation*}
where $\setmin_\tme $ is a random nonempty set. Then one of the arms in $\mathcal{I}_{\tme-1}$ is pulled at time $\tme$ if $\rv=1$. Every time one of the arms with the smallest number of pulls is pulled, either $\marm$ goes up by 1, or the cardinality of the set $\mathcal{I}_{\tme-1}$ is is decreased by 1. Since that cardinality cannot exceed $\numarm$, we see that, if one of the arms with the smallest number of pulls is pulled $\numarm$ times in a row, then the smallest number of times an arm is pulled goes up at least by 1. Therefore,
\begin{equation} \label{eq:LB}
\marm \geq  \left\lfloor  \frac{1}{\numarm} \sum_{d=1}^{\tme}J_{d} \right\rfloor
\geq    \frac{1}{\numarm} \sum_{d=1}^{\tme}J_{d}-1, \tme=1,2,\ldots. 
\end{equation}
Denote~$S_\tme=\sum_{d=1}^{\tme}J_{d},~\tme=1,2, \ldots$. Note that
\begin{equation*}
\expec (S_\tme)=\frac{1}{\numarm-1}\sum_{d=1}^{\tme}\varepsilon_{d}\to\infty
\end{equation*}
as $\tme \to\infty$ as the step-size sequence is not summable. Further,
\begin{equation*}
\expec(S_\tme- \expec(S_\tme))^2=\sum_{d=1}^{\tme}\frac{\varepsilon_d}{\numarm-1}\left(1-\frac{\varepsilon_d}{\numarm-1}\right) \leq
\frac{1}{\numarm-1}\sum_{d=1}^{\tme}\varepsilon_{d} = \expec(S_{\tme}).
\end{equation*}
We claim that the strong law of large numbers 
\begin{equation} \label{e:strong.law}
  \frac{K-1}{\sum_{d=1}^{n}\vep_{d}}S_n \to 1 \ \text{with probability 1}
\end{equation}
holds. To see that, denote $s_n=\sum_{d=1}^{n}\vep_{d}, \,
n=1,2,\ldots$ and define
$$
m_\ell=\min\bigl\{ n=1,2,\ldots: \, s_n\geq \ell^2\bigr\}, \ \ell=1,2,\ldots.
$$
Note that 
$$
ES_{m_\ell}=\frac{s_{m_\ell}}{K-1}\geq \frac{\ell^2}{K-1}, \ \ {\rm
  Var}(S_{m_\ell})\leq \frac{s_{m_\ell}}{K-1}\leq \frac{\ell^2+1}{K-1}.
$$
By the Chebyshev inequality, for any $\delta>0$, 
$$
 P\left( \left| \frac{K-1}{s_{m_\ell}}S_{m_\ell} -1 \right|>\delta\right)
 \leq \frac{(K-1)(\ell^2+1)}{\delta^2\ell^4}.
 $$
Since this expression is summable in $\ell$, we conclude by first
Borel-Cantelli lemma   that \eqref{e:strong.law} holds along
the subsequence $(m_\ell)$. Next, for $n>m_1$ let $K_n$ be such that
$$
m_{K_n-1}<n\leq m_{K_n}.
$$
Then
$$
S_{m_{K_n-1}}<S_n\leq S_{m_{K_n}}, \ \ s_{m_{K_n-1}}<s_n\leq s_{m_{K_n}},
$$
so
$$
\frac{S_{m_{K_n-1}}}{ s_{m_{K_n-1}}}\frac { s_{m_{K_n-1}}}{s_{m_{K_n}}}
\leq 
\frac{S_{m_{K_n-1}}}{ s_{m_{K_n-1}}}\frac { s_{m_{K_n-1}}}{s_n}
\leq \frac{S_n}{s_n}\leq \frac{S_{m_{K_n}}}{ s_{m_{K_n}}}\frac {
  s_{m_{K_n}}}{s_n}\leq \frac{S_{m_{K_n}}}{ s_{m_{K_n}}}\frac {
  s_{m_{K_n}}}{ s_{m_{K_n-1}}}.
$$
Since
$$
\frac {s_{m_{K_n}}}{ s_{m_{K_n-1}}}\leq \frac{K_n^2+1}{(K_k-1)^2}\to 1
$$
as $n\to\infty$, the convergence in \eqref{e:strong.law} holds along
all positive integers. It follows from (\ref{eq:LB}) and  (\ref{e:strong.law}) that for every
$\xi > \numarm (\numarm-1)$ for all $\tme$ large enough each arm will
be pulled at least~$\frac{1}{\xi}\sum_{d=1}^{\tme}
\varepsilon_{d}~\text{times}$. 
\end{proof}

\vspace{0.2in}

\begin{proof}[\textbf{Proof of Theorem~\ref{thm:Idx}}]
For the sake of clarity and exposition, let~$n = N_k(\tme-1)$, $m =
m(t-1)$, $x_1,\ldots, x_n$ the
rewards from arm $k$ and 
 $x_{(1:n)}\geq x_{(2:n)},\geq \ldots \geq x_{(n:n)}$ are the same 
rewards from the largest to the smallest. It is
clear that $\apol_{\arm}(\tme)$ is one these ordered rewards. For a
set $A\subseteq
\{1,\ldots,n\}$ of cardinality $m$ we have
$$
\max_{j\in A}x_j = x_{(L(A):n)},
$$
where
$$
 L(A) =\min\bigl\{  j=1, \ldots, n:\ \text{there is $j^\prime \in A$ 
   such that} \ x_{j^\prime} = x_{(j:n)}\bigr\}.
 $$
We  break the ties and make one-to-one
correspondence between an order statistic and the corresponding
observation in an arbitrary way. Note that 
$$
\text{there are exactly ${ n-i\choose m-1}$ sets $A$ with $L(A)=i$.}
$$
Therefore, 
\begin{align*}
\apol_{\arm}(\tme)= \mathcal{O}_{\arm, \tme-1}(l),~\text{where}~
l=\min\left\{  d\geq 1:\ \sum_{i=1}^d {n-i\choose m-1} \geq \frac12
     {n \choose m}    \right\}. \notag 
\end{align*}
We have by elementary combinatorics,
\begin{equation*}
\sum_{i=1}^d {n-i\choose m-1} =  {n\choose m} - {n-d\choose m}.
\end{equation*}
Therefore, we can write
\begin{equation*}
l =\min\left\{  d\geq 1:\   {n-d\choose m}\leq \frac12
     {n\choose m}    \right\}. 
\end{equation*}
Furthermore,
\begin{align*}
&\frac{ {n-d\choose m}}{ {n\choose m}} = \frac{(n-d)(n-d-1)\cdots (n-d-m+1)}{
                 n(n-1)\cdots (n-m+1)} \leq \left(\frac{n-d}{n}\right)^m,
\end{align*}                 
implying that  $l \leq \Big \lceil n\bigl( 1-2^{-1/m}\bigr) \Big \rceil$. 
Since $1-2^{-1/m}\leq (\log 2)/m$, we can see that
$l\ll \lceil \frac{2n}{m} \rceil$, and so
$$
\mathcal{O}_{\arm,\tme-1}\Big(\Big\lceil \frac{ 2
  N_k(t-1)}{m(t-1)} \Big\rceil \Big) \leq \apol_{\arm}(\tme).
$$

For an upper bound, notice that for large $n$  we
have, for some $\rho_n\downarrow 0$ (that may change from appearance to
appearance), by Stirling's formula, uniformly in $d$ in a bounded
range, 
\begin{align*}
&\frac{ {n-d\choose m}}{ {n\choose m}} =  \frac{(n-m)!}{(n-m-d)!}\frac{(n-d)!}{n!}\\
\geq
& (1-\rho_n)\frac{(n-m)^{n-m}}{(n-m-d)^{n-m-d}}\frac{(n-d)^{n-d}}{n^n}
       \frac{\sqrt{n-m}}{\sqrt{n-m-d}}\frac{\sqrt{n-d}}{\sqrt{n}}\\
\geq & (1-\rho_n)\left(1-\frac{m}{n-d}\right)^d  \geq
       (1-\rho_n)\exp\left\{-dC\frac{m}{n-d}\right\}
\end{align*}
for $C>0$ such that $1-x\geq e^{-Cx}$ for $0\leq x\leq
1/2$. Therefore, for large $n$, 
\begin{equation*}
l C\frac{m}{n-l}\geq \log(3/2)\,,
\end{equation*}
and since $m\leq n/2$,
$$
l\geq \tau:=\frac{2\log (3/2)}{2C}. 
$$
This gives us the upper bound
$$
 \apol_{\arm}(\tme)  \leq \postat\Big(\Big\lceil \frac{ \tau
   N_k(t-1)}{m(t-1)} \Big\rceil \Big), 
$$
and the result follows.
\end{proof}

\vspace{0.2in}

\begin{proof}[\textbf{Proof of Theorem \ref{thm:PVR}}]
Clearly, it is enough to prove the lower bound
\begin{equation} \label{eq:LPol}
  \liminf_{\tme \to\infty} \frac{V_{\tme}(\pi)}{\expec \Big[
    \max_{n\leq \tme}\reward^{(\optarm)}_n\Big]}\geq 1. 
\end{equation}
For $g \geq 1$ consider the event,
\begin{equation*}
A_g=\Big\{ W_{\optarm}(\tme) \leq \ipolkt~\text{for some $\tme > g$ and some $\arm \in \setarm / \{\optarm\}$}  \Big\}.
\end{equation*}
By Theorem ~\ref{thm:EThm1} we know that $\mathbb{P}(A_g)\to 0$ as $g \rightarrow \infty$. For $\tme >g$ we have
\begin{align*}
\val(\pi) &\geq \expec \Big[ \max_{g<n\leq\tme} \reward^{(\optarm)}_n \boldsymbol{1} \Big( B_n \cap A^{\mathrm{c}}_g \Big)  \Big] \\
&\geq \expec \Big[ \max_{g<n\leq\tme} \reward^{(\optarm)}_n   \boldsymbol{1}  (B_n) \Big] - \expec \Big[ \max_{g<n\leq \tme} \reward^{(\optarm)}_n \boldsymbol{1}  (B_n \cap A_g)  \Big] \\
&\geq   \expec \Big[ \max_{g<n\leq\tme} \reward^{(\optarm)}_n
   \boldsymbol{1}  (B_n) \Big] -\expec \Big[ \max_{g<n\leq\tme}  \reward^{(\optarm)}_n \boldsymbol{1}  (A_g)  \Big], 
\end{align*}
where $B_n$ is the event that the arm pulled at time $n$ is the arm with the highest index. Letting $K_{g+1,\tme}$ be the number of times between $g+1$ and $\tme$ that the arm with the highest index is not pulled, we have
\begin{equation*}
  \expec \Big[ \max_{g<n\leq\tme} \reward^{(\optarm)}_n
   \boldsymbol{1}  (B_n) \Big] 
= \expec \left[ \max_{j=1,\ldots,\tme-g-K_{\tme+1,\tme}}\reward^{(\optarm)}_j\right],
\end{equation*}
while by \eqref{eq:SupPA}, 
\begin{equation} \label{eq:Kcont}
\lim_{\tme \to \infty}\frac{\expec \Big[ \max_{j=1,\ldots,\tme-g-K_{g+1,\tme}}\reward^{(\optarm)}_j \Big]}{\expec \Big[ \max_{j=1,\ldots,\tme}\reward^{(\optarm)}_j \Big]}=1.
\end{equation}
To ensure~(\ref{eq:Kcont}), one needs to control the
size of $K_{g+1,t}$. Such control is provided by the fact that the
sequence $(\epsilon_t)$ converges to 0. To see this, for every $0<\delta<1$
\[
P\bigl(K_{g+1,t}>\delta t\bigr) \leq \frac{E K_{g+1,t}}{\delta t}
\leq \frac{\sum_{j=1}^t \epsilon_j/K}{\delta t} \to 0,
\]
as $t\to\infty$. Therefore, for any such $\delta$,
\begin{align*}
  &E\max_{j=1,\ldots, t-g-K_{g+1,t}}X_j^{(i^*)} \geq
  E\left[ \max_{j=1,\ldots, t-g-K_{g+1,t}}X_j^{(i^*)} \one\bigl(
    K_{g+1,t}\leq \delta t\bigr)\right]\\
 \geq &P\bigl(K_{g+1,t}\leq \delta t\bigr) E\max_{j=1,\ldots,
  (1-\delta)t-g }X_j^{(i^*)} \sim a_{i^*}^{1/\lambda_{i^*}}\Gamma\bigl(
        1-1/\lambda_{i^*}\bigr) \bigl((1-\delta)t\bigr) ^{1/\lambda_{i^*}}        
 \end{align*}
 as $t\to\infty$ by~\eqref{eq:SupPA}. Therefore,
\[
 \liminf_{t\to\infty} \frac{E\max_{j=1,\ldots,
     t-g-K_{g+1,t}}X_j^{(i^*)}}{E\max_{j=1,\ldots, t}X_j^{(i^*)}} 
\geq (1-\delta) ^{1/\lambda_{i^*}}. 
\]
Since this is true for all $0<\delta<1$, we obtain
\[
 \liminf_{t\to\infty} \frac{E\max_{j=1,\ldots,
     t-g-K_{g+1,t}}X_j^{(i^*)}}{E\max_{j=1,\ldots, t}X_j^{(i^*)}} 
\geq 1. 
\]
Furthermore,
\[
\frac{E \max_{j=1,\ldots,
     t-g-K_{g+1,t}}X_j^{(i^*)}}{E \max_{j=1,\ldots, t}X_j^{(i^*)}} \leq 1.
\] 
\noindent
This ensures that~$K_{g+1,t}$ behaves nicely. Therefore, \eqref{eq:LPol} will follow once we show that
\begin{equation} \label{eq:Smm}
\lim_{g\to\infty}\limsup_{\tme\to\infty} \tme^{-1/\lambda_{\optarm}}
\expec
\Big[ \max_{n\leq\tme}\reward^{(\optarm)}_n \boldsymbol{1}(A_g) \Big]=0.
\end{equation}
To this end, choose $1<\theta<\lambda_{\optarm}$, and note that
\begin{equation*}
\expec\Big[\max_{n\leq \tme}\reward^{(\optarm)}_n \boldsymbol{1}(A_m)
\Big] \leq \Big\{ \expec \Big[ \max_{n\leq \tme} \Big(\reward^{({\optarm})}_n \Big)^\theta\Big] \Big\}^{1/\theta} \Big(P(A_g)\Big)^{(\theta-1)/\theta}.
\end{equation*}
Replacing~$\lambda_{\optarm}$ by $\lambda_{\optarm}/\theta>1$, we have
by \eqref{eq:SupPA},
$$
\expec \Big[ \max_{n\leq \tme} \Big(\reward^{({\optarm})}_n
\Big)^\theta\Big] \leq s(\theta,\lambda_{\optarm}) \tme^{-\theta/\lambda_{\optarm}}
$$
for some $s(\theta,\lambda_{\optarm})$  finite positive constant
depending only on $\theta$ and $\lambda_{\optarm}$. Therefore, 
\begin{equation*}
\expec \left[ \max_{n\leq \tme}\reward^{(\optarm)}_n \boldsymbol{1}(A_g) \right] \leq
s(\theta,\lambda_{\optarm})^{1/\theta} \tme^{-1/\lambda_{\optarm}} \bigl(
\mathbb{P}(A_g)\bigr)^{(\theta-1)/\theta}. 
\end{equation*}
Since $\mathbb{P}(A_g)\to 0$ as $g\to\infty$,  \eqref{eq:Smm} follows.
\end{proof}

\vspace{0.2in}

\begin{proof}[\textbf{Proof of Theorem~\ref{thm:PSS}}]
Denote
\begin{equation*}
M_*=\sup\bigl\{ \tme \geq 1:\, W_{\optarm}(\tme)\leq \ipolkt \  \text{for some $k \neq i^*$} \bigr\}\,.
\end{equation*}
It follows from Theorem \ref{thm:EThm1} (which holds for
polynomial-like arms)  that $M_*<\infty$ a.s.  For $m\leq n \leq
\tme$, let $K_{m,n}$ denote the number of times between $m$ and $n$
that the arm with the highest index is not pulled. We have 
\begin{equation*}
V_n(\pi) \geq \E\left[  \max_{j=1,\ldots,
    n-M_*-K_{M_*+1,n}}X_j^{(\optarm)}\right].
\end{equation*}
Choose $1-\alpha<\theta<1-1/\lambda_{\optarm}$. We have 
\begin{align*} 
V_n(\pi) \geq& \E\left[  \max_{j=1,\ldots,
    n-[n^\theta]-K_{[n^\theta]+1,n}}X_j^{(\optarm)}\one\bigl(M_*\leq
  n^\theta\bigr)\right] \\
  =&\E\left[  \max_{j=1,\ldots,
     n-[n^\theta]-K_{[n^\theta]+1,n}}X_j^{(\optarm)} \right]
     - \E\left[  \max_{j=1,\ldots,
    n-[n^\theta]-K_{[n^\theta]+1,n}}X_j^{(\optarm)}\one\bigl(M_*>
  n^\theta\bigr)\right]. \notag 
\end{align*}
Note that
\begin{align*}
&\E\left[  \max_{j=1,\ldots,
  n-[n^\theta]-K_{[n^\theta]+1,n}}X_j^{(\optarm)} \right]
  \geq
\E\left[  \max_{j=1,\ldots,
  n-2[n^\theta]}X_j^{(\optarm)} \one\bigl( K_{1,n}\leq n^\theta\bigr)\right]
  \\
 \geq & \E\left[  \max_{j=1,\ldots,
  n-2[n^\theta]}X_j^{(\optarm)}\right]
-\E\left[  \max_{j=1,\ldots,
  n}X_j^{(\optarm)} \one\bigl( K_{1,n}> n^\theta\bigr)\right]. 
\end{align*}
Using the fact that $\beta_{\optarm} > 1/\lambda_{\optarm}$ we have by
Theorem 1 in~\cite{CV14},   for large $n$,  
\begin{align*}
  &\E\left[  \max_{j=1,\ldots,   n-2[n^\theta]}X_j^{(\optarm)}\right]\\
  \geq&
  \bigl((n-n^\theta)a_{\optarm}\bigr)^{1/\lambda_{\optarm}}\Gamma(1-1/\lambda_{\optarm})+o(1)
  \\
  \geq &(n a_{\optarm}) ^{1/\lambda_{\optarm}}\Gamma(1-1/\lambda_{\optarm})
     -   a_{\optarm}^{1/\lambda_{\optarm}}\Gamma(1-1/\lambda_{\optarm})
     \bigl(n^{1/\lambda_{\optarm}}-(n-n^\theta)^{1/\lambda_{\optarm}}\bigr)
     +o(1)\\
  =&(n a_{\optarm}) ^{1/\lambda_{\optarm}}\Gamma(1-1/\lambda_{\optarm}) +o(1)\\
     =& \E\left[  \max_{j=1,\ldots,   n}X_j^{(\optarm)}\right] +o(1),
\end{align*}
where the penultimate step is due to the choice of $\theta$. Therefore, result will follow once we prove that 
\begin{equation} \label{eq:ELR}
\lim_{n\to\infty} \E\left[  \max_{j=1,\ldots,
    n}X_j^{(\optarm)}\one\bigl(M_*> n^\theta\bigr)\right]
= \lim_{n\to\infty} \E\left[  \max_{j=1,\ldots,
    n}X_j^{(\optarm)} \one\bigl( K_{1,n}> n^\theta\bigr)\right]=0.
\end{equation} 
\noindent
Clearly,
\begin{align*}
&\E\left[  \max_{j=1,\ldots,
  n}X_j^{(\optarm)} \one\bigl( K_{1,n}> n^\theta\bigr)\right]  \\
  =& \E\left[  \max_{j=1,\ldots,
                 n}X_j^{(\optarm)}\right] \mathbb{P} \bigl( K_{1,n}>
     n^\theta\bigr).
\end{align*}
Since $\theta>1-\alpha$, an elementary exponential Markov inequality
shows that $\mathbb{P} \bigl( K_{1,n}>     n^\theta\bigr)$ is
exponentially small. Using \eqref{eq:SupPA}, we obtain one of the two
statements in \eqref{eq:ELR}. Similarly, 
\begin{align} \label{eq:St1}
&\E\left[  \max_{j=1,\ldots,
  n}X_j^{(\optarm)} \one\bigl( M_*> n^\theta\bigr)\right]  
  = \E\left[  \max_{j=1,\ldots,
                 n}X_j^{(\optarm)}\right] P \bigl( M_*> n^\theta\bigr)\,.
\end{align}
Next we estimate the probability in the right hand side above. Note
that this probability does not change if we apply the same monotone
increasing function to all rewards. Taking the logarithm of the
rewards makes the reward distribution exponential-like, i.e.,  satisfy~\eqref{eq:TE}. 
Let $0<\delta<(\min_{i\not= i^*}\lambda_i-\lambda_{\optarm})/\lambda_{\optarm}$. We have
\begin{align} \label{eq:St2} 
  \mathbb{P} \bigl( M_*> n^\theta\bigr) \leq & \mathbb{P} \bigl( W_{\optarm}(d)\leq
  (1+\delta)^{-1/2} \lambda_{\optarm}^{-1} \log m(d-1) \  \text{for some
   $d>n^\theta$}\bigr)\\
  +\sum_{i\not= i^*}  & \mathbb{P} \bigl( W_i(d)\geq
  (1+\delta)^{1/2} \lambda_i^{-1} \log m(d-1) \  \text{for some
   $d>n^\theta$}\bigr)\,. \notag 
  \end{align}
Next,  for $a>(\numarm -1)K$,
\begin{align*}
& \mathbb{P} \bigl( W_{\optarm}(d)\leq
  (1+\delta)^{-1/2} \lambda_{\optarm}^{-1} \log m(d-1) \  \text{for some
                 $d>n^\theta$}\bigr) \\
  \leq&    \sum_{d>n^\theta}  \mathbb{P} \left( m(d-1)\leq  \frac{1}{a} \sum_{l=1}^{d-1} \vep_l\right)\\
+&      \sum_{d>n^\theta}    \mathbb{P} \left( m(d-1)>  \frac{1}{a} \sum_{l=1}^{d-1}
  \vep_l, \  W_{\optarm}(d)\leq
  (1+\delta)^{-1/2} \lambda_{\optarm}^{-1}  \log m(d-1)\right)\,.  
\end{align*}
Since $m(d-1)\geq K_{1,d-1}/(K-1)$, another application of the exponential
Markov inequality shows that $\mathbb{P} \left( m(d-1)\leq
  \sum_{l=1}^{d-1} \vep_l/a\right)$ decreases exponentially fast, hence
the sum 
\begin{equation*}
 \sum_{d>n^\theta}  \mathbb{P} \left( m(d-1)\leq  \frac{1}{a}
   \sum_{l=1}^{d-1} \vep_l\right) 
\end{equation*}
is an exponetially fast decreasing function of $n$. Furthermore, 
 \begin{align*}
 &\mathbb{P} \left( m(d-1)>  \frac{1}{a} \sum_{l=1}^{d-1}
  \vep_l, \  W_{\optarm}(d)\leq
   (1+\delta)^{-1/2} \lambda_{\optarm}^{-1}  \log m(d-1)\right) \\
\leq &\sum_{ {\sum_{l=1}^{d-1} \vep_l/a<j_1\leq d}\atop {j_1\leq
       j_2\leq d}} 
\mathbb{P} \left( \lceil j_2/j_1\rceil^{\rm th} \ \text{order statistic
       in} \ X_1^{(i^*)}, \ldots,  X_{j_2}^{(i^*)} \leq
   (1+\delta)^{-1/2} \lambda_{\optarm}^{-1}  \log j_1\right).
 \end{align*}
 The latter sum is a sum of binomial probabilities and the exponential
 markov inequality shows that it also decays exponetially fast with
 $d$, hence the sum
 $$
 \sum_{d>n^\theta}    \mathbb{P} \left( m(d-1)>  \frac{1}{a} \sum_{l=1}^{d-1}
  \vep_l, \  W_{\optarm}(d)\leq
  (1+\delta)^{-1/2} \lambda_{\optarm}^{-1}  \log m(d-1)\right)
$$
decays exponentially fast with $n$.    It follows that
\begin{equation} \label{eq:Arg1}
 \lim_{n\to\infty}\left\{ \E\left[  \max_{j=1,\ldots,
                 n}X_j^{(\optarm)}\right] \mathbb{P} \bigl( W_{\optarm}(d)\leq
  (1+\delta)^{-1/2} \lambda_1^{-1} \log m(d) \  \text{for some
    $d>n^\theta$}\bigr)\right\}=0\,.
 \end{equation}
\noindent
In an analogous way we can show that for any 
  that for any $i\not= i^*$, 
\begin{equation} \label{eq:Arg2}
 \lim_{n\to\infty}\left\{ \E\left[  \max_{j=1,\ldots,
                 n}X_j^{(\optarm)}\right] \mathbb{P} \bigl( W_i(d)\geq
  (1+\delta)^{1/2} \lambda_{\optarm}^{-1} \log m(d) \  \text{for some
    $d>n^\theta$}\bigr)\right\}=0,
 \end{equation}
and the remaining statement in \eqref{eq:ELR} follows from \eqref{eq:St1}, \eqref{eq:St2}, \eqref{eq:Arg1} and \eqref{eq:Arg2}.
\end{proof}

\vspace{0.1in}

\noindent
\textbf{Remark}: One can prove Theorem~\ref{thm:EVER} using the same
arguments as in Theorem \ref{thm:PSS}, thereby establishing the
vanishing extremal regret in the strong sense in case of exponential-like arms.

\vspace{0.2in}

\begin{proof}[\textbf{Proof of Theorem~\ref{thm:MAS}}]
Note that one can switch from exponential-like arms to polynomial-like arms by exponentiation of the former, and switch back by taking the logarithm of the latter. The result is invariant under monotone transformation of the rewards, and the theorem holds for both exponential-like and polynomial-like arms. We establish the result for exponential-like arms below. \\
Let $a>(K-1)K$, $b > 0$ and $A>\kappa+2$. Denote
$\upsilon_{d-1}=\sum_{l=1}^{d-1} \epsilon_l/a$. Using
\eqref{e:ld.upper.exp}, we have for  $k\not= i^*$, for some~$c>0$ 
\begin{align}
&\mathbb{P}\left(m(d-1) \geq  \frac1a\sum_{l=1}^{d-1} \epsilon_l, \widetilde{\ipol}_{\arm}(d) \geq \frac{1}{\lambda_k} \log a_{\arm} + \frac{1}{\lambda_k} \log h(m(d-1)) + b \right) \nonumber \\
  \leq &\sum_{m \geq \upsilon_{d-1}} \sum_{j = m}^{d-1} \mathbb{P}
   \left( \text{$\lceil j/h(m)\rceil$th order stastitic out of $j$}\ \geq \frac{1}{\lambda_k} \log a_{\arm} + \frac{1}{\lambda_k} \log h(m) + b \right) \nonumber \\
\leq &\sum_{m \geq \upsilon_{d-1}} \sum_{j = m}^{d-1}  \exp(-cj/h(m)) \leq \sum_{m \geq \upsilon_{d-1}}  \frac{\exp(-cm/h(m))}{1 - \exp(-c/h(m))} \nonumber \\
\leq& \frac{1}{c} \sum_{m \geq \upsilon_{d-1}} h(m) \exp(-cm/h(m)) \nonumber \\ 
\leq& \frac{1}{c} \sum_{m \geq \upsilon_{d-1}} h(m) \exp(-A \log
      m) \ \ (\text{because $h(x)=o(x/\log x)$} )\nonumber \\
\leq& \frac{1}{c} \sum_{m \geq \upsilon_{d-1}} m^{A-1}  = O\left(\left( \sum_{l=1}^{d-1} \epsilon_l
   \right)^{-(A-2)}\right)
\end{align}
for all~$d$ large enough. 
\noindent
Using the condition on probabilities~$(\epsilon_t)$ and 
the first Borel-Cantelli lemma; see e.g.,~\cite{Dur19}, we see that for
any~$b>0$ 
\begin{align*}
 \mathbb{P}\Bigl(m(d-1) \geq  \frac1a\sum_{l=1}^{d-1} \epsilon_l,\, 
  \widetilde{\ipol}_{\arm}(d) \geq& \frac{1}{\lambda_k} \log a_{\arm}  \\
+&
  \frac{1}{\lambda_k} \log h(m(d-1)) + b \ \text{for infinitely many
   $d$}\Bigr)  = 0.
\end{align*}
Since~$a > \numarm-1$, by Theorem~\ref{thm:SS}, for any $b>0$
 \begin{align*}
 \mathbb{P}\Bigl( 
  \widetilde{\ipol}_{\arm}(d) \geq \frac{1}{\lambda_k} \log a_{\arm}  
+
  \frac{1}{\lambda_k} \log h(m(d-1)) + b \ \text{for infinitely many
   $d$}\Bigr)  = 0.
\end{align*}
An analogous argument shows that for any $b>0$
\begin{align*}
 \mathbb{P}\Bigl( 
  \widetilde{\ipol}_{\arm}(d) \leq \frac{1}{\lambda_k} \log a_{\arm}  
+
  \frac{1}{\lambda_k} \log h(m(d-1)) - b \ \text{for infinitely many
   $d$}\Bigr)  = 0, 
\end{align*}
and the result follows.
\end{proof}

\vspace{0.2in}

\begin{proof}[\textbf{Proof of Theorem \ref{thm:EThm1}}]
 The proof closely follows the proof of Theorem~\ref{thm:MAS}, and is established below for exponential-like rewards. Because exponential-like and polynomial-like arms are related via
a monotone transformation, it is enough to prove the statement for
exponential-like arms.  \\ 
First, we establish that for any $a>(K-1)K$ and $\delta>0$,
\[
P\Big(m(n) \geq \frac{1}{a} \sum_{d=1}^n \epsilon_d, W_i(n) \geq (1+\delta) \lambda_i^{-1} \log m(n)~\text{for infinitely many}~n \Big) = 0.
\]
The above probability can be bounded 
as follows   
\[
P\Big(m(n) \geq \frac{1}{a} \sum_{d=1}^n \epsilon_d, W_i(n) \geq (1+\delta) \lambda_i^{-1} \log m(n) \Big) \leq \sum_{m \geq \frac{1}{a} \sum_{d=1}^n \epsilon_d} \sum_{k = m}^n P\Big( \mathcal{O}_{i,k} \Big( \Big\lfloor \frac{k}{m} \Big\rfloor \Big) \geq (1+\delta) \lambda_i^{-1} \log m \Big). \] 
The double summation on the right-hand side is the probability that a Binomial random variable with $k$ trials and the probability for success $a_i~m^{-(1+\delta)}$, takes a value at least $k/m$. Using the exponential Markov inequality for the Binomial random variable $X\sim B(k,p)$, for any $\theta>0$,
\[
  P(X>m)\leq e^{-\theta m} \left(e^\theta p + 1-p\right)^k,
\]
with $\theta=(1+\delta)\log m$ gives us the upper bound of 
\[
 \exp\left\{-k^\delta/2\right\}.
\]
That is we have for~$0 < A < \infty$,
\[
P\Big(m(n) \geq \frac{1}{a} \sum_{d=1}^n \epsilon_d, W_i(n) \geq (1+\delta) \lambda_i^{-1} \log m(n) \Big) \leq \exp\Big\{ -A \Big(\sum_{d=1}^n \epsilon_d \Big)^{\delta} \Big\}.
\]
Now using Borel-Cantelli lemma, we have that for any sub-optimal arm $i$ and any
$0<\delta<1$, w.p.1,
\[
W_i(n)\leq (1+\delta)\lambda_i^{-1}\log m(n)
\]
for all $n$ large enough. This is because the event
\[
\bigl\{ m(n)\geq \frac1a\sum_{d=1}^n \epsilon_d\ \text{for infinitely many}~n\bigr\}
\]
has probability~$1$. 
Since the optimal $\lambda_{i^*}$ is strictly smaller than the next best $\lambda_i$, we can find $0<\delta<1$ so that 
\[
(1-\delta)\lambda_{i^*}^{-1}>(1+\delta)
\lambda_i^{-1},
\]
for all sub-optimal $i$. \\
\noindent
Next, we establish that for any $a>(K-1)K$
\[
P\Big(m(n) \geq \frac{1}{a} \sum_{d=1}^n \epsilon_d, W_i(n) < (1-\delta) \lambda_i^{-1} \log m(n)~\text{for infinitely many}~n \Big) = 0.
\]
The above probability can similarly be upper bounded by 
 \[
  \sum_{m\geq \sum_{d=1}^n\vep_d/a}\sum_{k=m}^n P\Big( \mathcal{O}_{i,k} \Big( \Big\lfloor \frac{k}{m} \Big\rfloor \Big) < (1-\delta) \lambda_i^{-1} \log m \Big).
  \]
This probability is the probability that a Binomial random variable with $k$ trials and the probability for success $a_im^{-(1-\delta)}$, takes a value smaller than $k/m$. Again using the exponential Markov inequality 
for the Binomial random variable $X\sim B(k,p)$: for any $\theta>0$,
\[
  P(X<m)\leq e^{\theta m} \left(e^{-\theta} p + 1-p\right)^k.
\]
With $\theta=(1-\delta)\log m$, we similarly obtain for all $0<\delta<1$, 
\[
  W_{i^*}(n)\geq (1-\delta)\lambda_{i^*}^{-1} \log m(n)
\]
for all $n$ large enough w.p.1, and the result holds. 
\end{proof}

\bibliography{references}
\bibliographystyle{plainnat}

\end{document}